\documentclass[10pt,twocolumn,letterpaper]{article}

\usepackage[pagenumbers]{cvpr} 

\usepackage{multirow}
\usepackage{colortbl}
\definecolor{lightgray}{gray}{0.92}
\definecolor{cvprblue}{rgb}{0.21,0.49,0.74}
\usepackage[pagebackref,breaklinks,colorlinks,citecolor=cvprblue]{hyperref}

\def\paperID{382} 
\def\confName{CVPR}
\def\confYear{2025}

\title{\textsc{StableV2V}: Stablizing Shape Consistency in Video-to-Video Editing}

\author{%
  Chang Liu$^1$, Rui Li$^1$, Kaidong Zhang$^1$, Yunwei Lan$^1$, Dong Liu$^1$\thanks{Corresponding author.} \\
  $^1$University of Science and Technology of China \\
  \texttt{\{lc980413, liruid, richu, ywlan\}@mail.ustc.edu.cn,}\\
  \texttt{dongeliu@ustc.edu.cn}\\
  Project page: \url{https://alonzoleeeooo.github.io/StableV2V}
}

\begin{document}
\twocolumn[{
 \renewcommand\twocolumn[1][]{#1}%
 \maketitle
 \begin{center}
  \centering
  \includegraphics[width=1.0\textwidth]{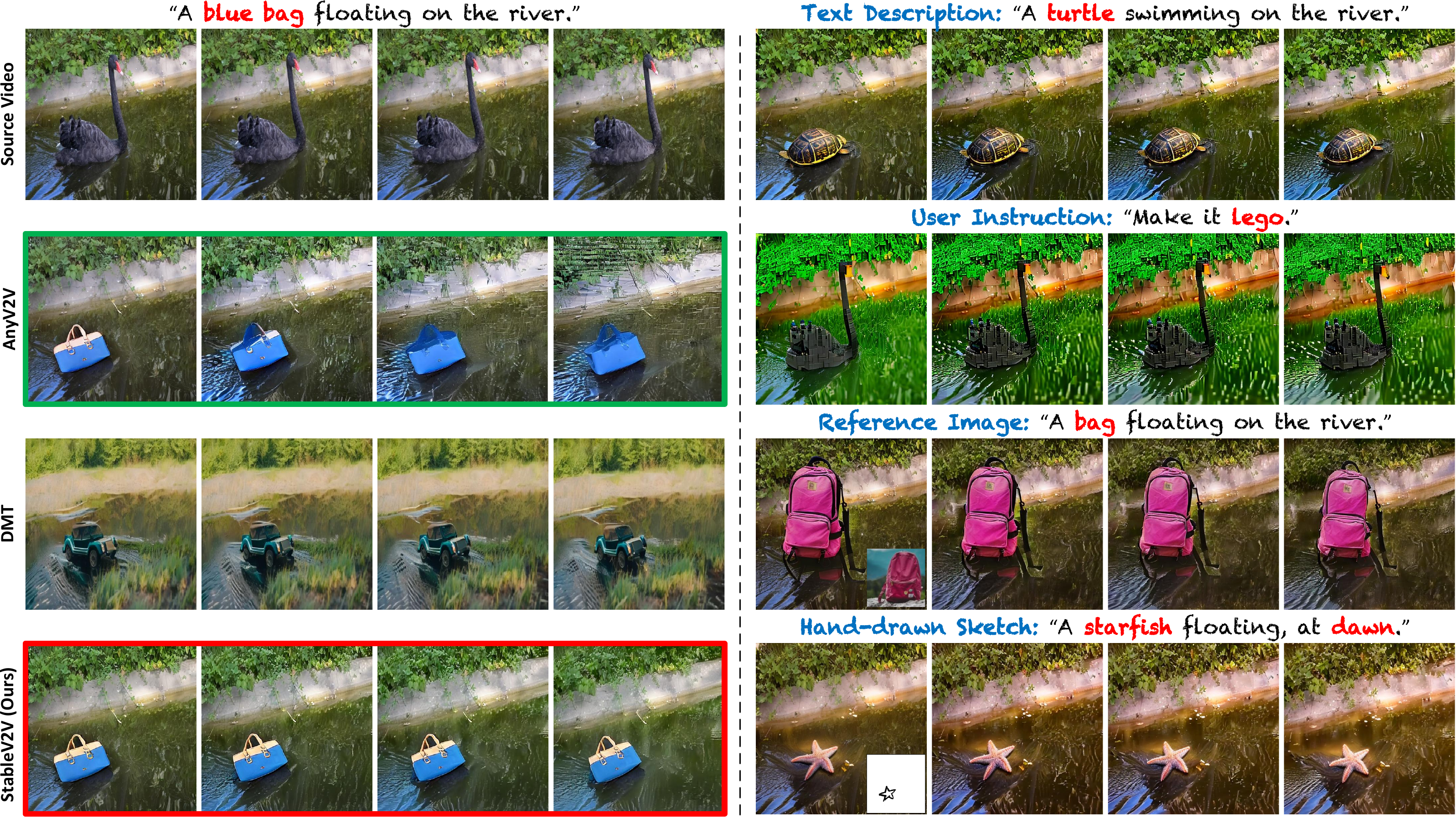}
  \vspace{-1.8em}
  \captionof{figure}{
  \textbf{Qualitative comparison (left) and results on different editing tasks by \textsc{StableV2V} (right).}
  Herein, we highlight the words that depict the main edited contents and the modalities of external prompts in red and blue, respectively, and present the visualizations of several prompts (i.e., reference image and hand-drawn sketch) at the right-bottom corner of the corresponding first edited frames.
  Notably, AnyV2V \cite{ku-etal-2024-anyv2v} uses the same first edited frames as ours, where both results are highlighted in green and red bounding boxes, respectively.
  }
  \label{fig: teasor}
 \end{center}}]

\begin{abstract}
Recent advancements of generative AI have significantly promoted content creation and editing, where prevailing studies further extend this exciting progress to video editing.
In doing so, these studies mainly transfer the inherent motion patterns from the source videos to the edited ones, where results with inferior consistency to user prompts are often observed, due to the lack of particular alignments between the delivered motions and edited contents.
To address this limitation, we present a shape-consistent video editing method, namely StableV2V, in this paper.
Our method decomposes the entire editing pipeline into several sequential procedures, where it edits the first video frame, then establishes an alignment between the delivered motions and user prompts, and eventually propagates the edited contents to all other frames based on such alignment.
Furthermore, we curate a testing benchmark, namely DAVIS-Edit, for a comprehensive evaluation of video editing, considering various types of prompts and difficulties.
Experimental results and analyses illustrate the outperforming performance, visual consistency, and inference efficiency of our method compared to existing state-of-the-art studies.\footnote{We open-source our codebase at \url{https://github.com/AlonzoLeeeooo/StableV2V}, and release the model weights and testing benchmark \textsc{DAVIS-Edit} at \url{https://huggingface.co/AlonzoLeeeooo/StableV2V} and \url{https://huggingface.co/datasets/AlonzoLeeeooo/DAVIS-Edit}, respectively.}

\end{abstract}

\vspace{-1em}
\section{Introduction}
\vspace{-0.2em}
Video editing aims to modify the source video contents according to user demands.
With the prosper of diffusion models \cite{ho-etal-2020-ddpm, rombach-etal-2022-stable-diffusion} that demonstrates superior generative capabilities, recent studies have adopted this astonishing technique for video editing, making it possible for end users to interact with various types of external prompts, e.g., text \cite{geyer-etal-2024-tokenflow, zhang-etal-2024-avid}, instruction \cite{wu-etal-2023-fairy, yuan-etal-2023-instructvideo}, image \cite{peruzzo-etal-2024-vase, xiao-etal-2023-fastcomposer}, sketches \cite{liu-etal-2024-sketchrefiner}, and etc.
They achieve significant success on this topic, bringing video editing to a prominent attractive research direction for the community of visual content generation.

To perform video editing, recent studies manage to transfer the motion patterns from the original video and adapt them to the editing process.
In doing so, prevailing methods can be categorized into four main types, i.e., DDIM inversion-, one-shot tuning-, learning-, and first-frame-based methods. 
Specifically, DDIM inversion-based methods \cite{geyer-etal-2024-tokenflow, danah-etal-2024-dmt} leverage DDIM inversion to store the motion patterns of videos in forms of latent features, which are then injected into the diffusion models when editing, thus enforcing the consistency between edited frames and the original ones.
One-shot tuning-based solutions \cite{jay-etal-2023-tuneavideo, liu-etal-2023-videop2p} aim to tailor the motion patterns of each video through learning video-specific model weights.
These two types of methods, however, often produce results that are inconsistent to the shapes that user prompts require, especially the ones with significant shape differences, e.g., the cases illustrated in Fig.~\ref{fig: teasor}.
Learning-based methods \cite{zi-etal-2024-cococo, zhang-etal-2024-avid, peruzzo-etal-2024-vase} provide a more general solution for video editing by fine-tuning temporal-enhanced diffusion models on large-scale video-text datasets \cite{webvid, panda70m}, but these studies are highly restricted due to their inpainting paradigms. 
They normally require mask annotations to precisely localize the edited regions, thus becoming tough for users to interact with.
Also, the inpainting paradigms limit them to regional editing scenarios, where the applications of global ones (e.g., video style transfer \cite{lai-etal-2018-learning}) are neglected. 
First-frame-based methods \cite{ku-etal-2024-anyv2v, ouyang-etal-2024-i2vedit} offer a more flexible solution for video editing, where this paradigm decomposes video editing into image editing and motion transfer, enabling the potentials to perform both global and local editing with the same solution.
Nevertheless, they suffer from similar limitations to the aforementioned studies due to their requirements of DDIM inversion \cite{ku-etal-2024-anyv2v} and video-specific tuning \cite{ouyang-etal-2024-i2vedit}.
Recently, DMT \cite{danah-etal-2024-dmt}, which proposes a space-time feature loss to constrain the motion consistency, serves as the most relevant study to address such misalignment, but even so, inferior condition-following ability and detail loss of backgrounds are often observed in its results like the ones in Fig. \ref{fig: teasor}, where effective paradigm is thus expected to ensure the consistency between delivered motions and user prompts.

Therefore in this paper, we propose \textsc{StableV2V} to perform video editing in a shape-consistent manner, with our method built based on the first-frame-based paradigm.
In doing so, our method performs video editing with three main components, i.e., Prompted First-frame Editor (PFE), Iterative Shape Aligner (ISA), and Conditional Image-to-video Generator (CIG).
PFE serves as the first-frame image editor that converts external prompts into edited contents, which are then propagated to other frames in later processes to construct the entire edited video.
To offer precise guidance that are well aligned with shapes required by user prompts, especially in scenarios that comprise complicated shape differences, we assume that the edited contents share the same motions with the ones of source video.
Based on the assumption, we propose ISA, which manages to iteratively propagate the average motions, shapes, and depths from core elements (e.g., main objects) of each original video frame to the edited one, resulting in the simulated optical flow and depth map of all edited frames, along with a shape-guided depth refinement network to further calibrate the obtained depth map and ensure its preciseness.
Eventually, we leverage the depth map as an intermediate vehicle to deliver precise motions from the source video, and utilize it to guide the image-to-video generation process of CIG, obtaining the final edited video.
Furthermore, we collect a testing benchmark based on DAVIS \cite{ponttuset-etal-2018-davis}, namely \textsc{DAVIS-Edit}, to conduct a comprehensive evaluation for text- and image-based video editing.
Experimental results compared to existing state-of-the-art studies demonstrate that \textsc{StableV2V} outperforms others from various perspectives, including visual quality, consistency, and inference efficiency.

\begin{figure*}[t!]
\centering
\includegraphics[width=1.0\linewidth, trim=0 0 0 0]{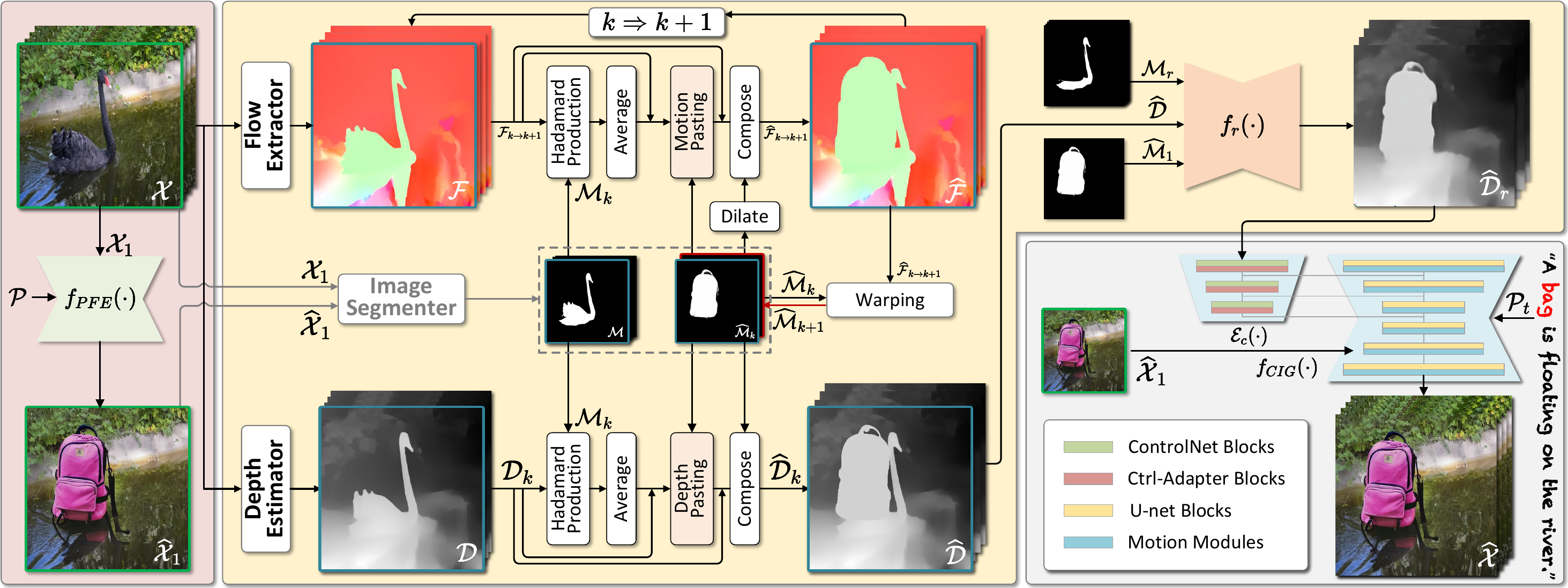}
\vspace{-2em}
\caption{
\textbf{Illustration of the overall pipeline of \textsc{StableV2V},} with three main components, i.e., Prompted First-frame Editor (PFE), Iterative Shape Aligner (ISA), and Conditional Image-to-video Generator (CIG), whose backgrounds are highlighted in red, yellow, and gray, respectively.
Herein, the green bounding boxes refer to the first video frames; the blue bounding boxes represent the $k$-th optical flow, segmentation mask, and depth map in ISA.
For simplicity, we only showcase the $k$-th to $k+1$-th iteration process of ISA in this figure.
}
\vspace{-1.8em}
\label{fig: pipeline}
\end{figure*}

\vspace{-0.5em}
\section{Related Works}
\noindent \textbf{Video Synthesis.}
Modeling the high-dimensional distribution of video data is a challenging task for video generation. 
Early-proposed methods \cite{tulyakov-etal-2018-mocogan} mainly address this problem via Generative Adversarial Network (GAN), but suffering from inferior visual quality and training instability.
Recent advancements of diffusion models \cite{ho-etal-2020-ddpm, rombach-etal-2022-stable-diffusion} have greatly promoted the development of various visual generation tasks, e.g., text-to-image and conditional generation \cite{kim-etal-2022-diffusionclip, liu-etal-2023-sdg, liu-etal-2024-lacon}, where this effective paradigm is also adopted for video generation \cite{wang-etal-2023-recipe, ma-etal-2024-latte}.
Particularly, existing studies leverage various model architectures upon the video modeling task, including U-net \cite{guo-etal-2023-animatediff} and Diffusion Transformer \cite{ma-etal-2024-latte}.
These studies demonstrate outstanding generative abilities in producing photo-realistic videos with text prompts, and serve as strong foundation models for a wide range of down-stream applications, e.g., text-to-video generation \cite{wang-etal-2023-lavie}, image-to-video generation \cite{hu-etal-2023-animateanyone, blattmann-etal-2023-svd, zhang-etal-2023-i2vgenxl, guo-etal-2024-i2vadapter} as well as video editing \cite{bai-etal-2024-uniedit, danah-etal-2024-dmt, geyer-etal-2024-tokenflow, cong-etal-2024-flatten, 
jay-etal-2023-tuneavideo, liu-etal-2023-videop2p, zi-etal-2024-cococo, ku-etal-2024-anyv2v}.

\noindent \textbf{Video Editing.}
Recently, the research direction of video editing has attracted great attention.
In performing this task, conventional works normally introduce external conditions to assist video editing, e.g., optical flow \cite{cong-etal-2024-flatten}, Neural Layered Atlas (NLA) \cite{omer-etal-2022-text2live, lee-etal-2023-shapeaware}, and etc., where limitations are usually observed due to the inherent problems of the used techniques.
With the prosper of diffusion models, such task is significantly facilitated by their strong generative abilities, where we summarize existing methods into four categories, i.e., DDIM inversion-, one-shot tuning-, learning-, and first-frame-based methods.
Specifically, DDIM inversion-based methods offer a way to represent the motion patterns of videos through inverted latent features, where these features are then utilized to enforce the temporal consistency in the generated video frames \cite{geyer-etal-2024-tokenflow}.
One-shot tuning-based methods \cite{jay-etal-2023-tuneavideo, liu-etal-2023-videop2p} mainly learn video-specific model weights to model the motion patterns, where diversified results can be then generated through adjusting the text prompts.
Learning-based methods \cite{zi-etal-2024-cococo, zhang-etal-2024-avid, peruzzo-etal-2024-vase} solve the task via training particular networks on large datasets, where they integrate motion modules into pre-trained image diffusion models \cite{rombach-etal-2022-stable-diffusion, yang-etal-2023-paint}, and optimize the enhanced model architectures with video-text data, enabling these networks to edit video contents in local regions.
First-frame-based methods \cite{ku-etal-2024-anyv2v, ouyang-etal-2024-i2vedit} start with editing the first video frame, and propagate the results to all other frames through transferring the motions from the source video.
Nevertheless, these studies obtain inferior performance since their delivered motions are inconsistent with user prompts.
AnyV2V \cite{ku-etal-2024-anyv2v} and DMT \cite{danah-etal-2024-dmt} are the most relevant studies to our method.
However, the former struggles to handle challenging scenarios with significant shape differences, and the latter presents inferior capability of background preservation, where all issues above motivate \textsc{StableV2V} in this paper.

\vspace{-0.6em}
\section{Methods}
\vspace{-0.2em}
\textsc{StableV2V} comprises three main components to perform video editing, i.e., Prompted First-frame Editor (PFE), Iterative Shape Aligner (ISA), and Conditional Image-to-video Generator (CIG), where the overall pipeline is shown in Fig. \ref{fig: pipeline}.
Given an input video $\mathcal{X} = \{ \mathcal{X}_1, \dots, \mathcal{X}_N \}$ with $N$ video frames in total, PFE edits the first video frame $\mathcal{X}_1$ into $\widehat{\mathcal{X}}_1$ according to an external prompt $\mathcal{P}$.
Then, ISA extracts the depth maps $\mathcal{D}$, optical flows $\mathcal{F}$, and segmentation masks $\mathcal{M}$ from $\mathcal{X}$, and simulates the depth maps $\widehat{\mathcal{D}}_r$ of edited video based on $\mathcal{D}$, $\mathcal{F}$, $\mathcal{M}$, and $\widehat{\mathcal{M}}_1$ of $\widehat{\mathcal{X}}_1$.
Eventually, CIG serves as a depth-guided image-to-video generator, and leverages $\widehat{\mathcal{D}}_r$ and $\widehat{\mathcal{X}}_1$ to produce the entire edited video $\widehat{\mathcal{X}}$, where the overall process of \textsc{StableV2V} is formulated by:
\vspace{-0.4em}
\begin{equation}
\vspace{-0.3em}
\small
    \widehat{\mathcal{X}} = f_{CIG} \left( f_{PFE} \left( \mathcal{X}_1, \mathcal{P} \right), f_{ISA} \left( \mathcal{D}, \mathcal{F}, \mathcal{M}, \widehat{\mathcal{M}}_1 \right) \right),
\end{equation}
where $f_{PFE} \left( \cdot \right)$, $f_{ISA} \left( \cdot \right)$, and $f_{CIG} \left( \cdot \right)$ denote PFE, ISA, and CIG, respectively.
In the following texts, we illustrate the details of each aforementioned component following the pipeline sequence of \textsc{StableV2V}.

\vspace{-0.3em}
\subsection{Prompted First-frame Editor}
\vspace{-0.2em}
Since \textsc{StableV2V} is built based on first-frame-based methods that decompose video editing into image editing and controlled image-to-video generation, the first step of \textsc{StableV2V} is to convert the external prompt into edited contents in the first video frame, with PFE serving as the core component in this step.
Given an input video $\mathcal{X} = \{ \mathcal{X}_1, \dots, \mathcal{X}_N \}$, we send its first frame $\mathcal{X}_1$ and the external prompt $\mathcal{P}$ into PFE, where we formulate this process by:
\vspace{-0.5em}
\begin{equation}
\vspace{-0.5em}
    \widehat{\mathcal{X}}_1 = f_{PFE} \left( \mathcal{X}_1, \mathcal{P} \right),
\end{equation}
where $\widehat{\mathcal{X}}_1$ refers to the first edited video frame of $\widehat{\mathcal{X}}$.
Herein, we consider various categories of prompt inputs $\mathcal{P}$, e.g., text descriptions, user instructions, reference images, and etc., where we adopt off-the-shelf image editors to process these prompts accordingly.
For example, we utilize text-guided editors, e.g., SD Inpaint \cite{rombach-etal-2022-stable-diffusion} and InstructPix2Pix \cite{brooks-etal-2023-instructpix2pix}, to process text inputs, and adopt models like Paint-by-Example \cite{yang-etal-2023-paint} to integrate reference image prompts.
Afterward in the subsequent processes, we build the alignment between motion controls and edited contents based on $\widehat{\mathcal{X}}_1$.

\vspace{-0.3em}
\subsection{Iterative Shape Aligner}
\vspace{-0.2em}
Once we obtain the first edited frame $\widehat{\mathcal{X}}_1$, the next step is to propagate the edited contents to the remaining video frames.
To conduct this step, we observe that existing studies often produce inferior results through directly propagation of motions from the source video, where the delivered motions in such case struggle to be consistent with contents that users expect, especially in the cases that user prompts may cause significant shape changes, as is shown in Fig. \ref{fig: teasor}, thereby leading to artifacts in the edited video.
Therefore, it is pivotal to propose an effective design to address such misalignment, so as to ensure the consistency in video editing.

In doing so, we propose ISA, which establishes the alignment between delivered motions and user prompts, and later offers precise guidance for CIG to produce the final video.
Specifically, we assume that \textit{the edited and original contents share the same \textbf{motion} and \textbf{depth} information}, and consider depth map as the intermediate media to deliver the motion information.
Based on the assumption, ISA sequentially simulates the motion and depth information of all edited video frames, and leverages an additional refinement network to obtain precise motion guidance for CIG.

\noindent \textbf{Motion Simulation.}
To simulate the motion information of the edited video, we use optical flows to represent its motions.
Given the source video input $\mathcal{X} = \{ \mathcal{X}_1, \dots, \mathcal{X}_N \}$ with $N$ frames, we utilize an off-the-shelf flow extractor (i.e., RAFT \cite{teed-etal-2020-raft}) to annotate the optical flows $\mathcal{F} = \{ \mathcal{F}_{1 \rightarrow 2}, \dots, \mathcal{F}_{N-1 \rightarrow N} \}$ from $\mathcal{X}$.
Besides, we use an image segmenter (i.e., SAM \cite{kirillov-etal-2023-sam}) to obtain the segmentation masks of all frames in $\mathcal{X}$, as well as the one of $\widehat{\mathcal{X}}_1$, resulting in $\mathcal{M} = \{ \mathcal{M}_1, \dots, \mathcal{M}_{N} \}$ and $\widehat{\mathcal{M}}_1$, respectively.
Considering that the edited contents and the original ones share the same motion information, we firstly compute the mean value of the $k$-th optical flow $\mathcal{F}_{k \rightarrow k+1}$ within $\mathcal{M}_k$ to represent the average motion, with the process formulated by:
\vspace{-0.8em}
\begin{equation} \label{eq: average-flow}
\vspace{-0.5em}
    \Bar{\mathcal{F}}_{k \rightarrow k+1} = \frac{1}{\mathcal{M}_k} \sum_{(i,j) \in \mathcal{M}_k} \mathcal{F}_{k \rightarrow k+1} \left( i, j \right),
\end{equation}
where $\left( i, j \right)$ represents the pixel at the $i$-th row and the $j$-th column of $\mathcal{M}_k$. Then, we simulate the flow within the regions of edited contents through performing the motion pasting operation on $\widehat{\mathcal{M}}_k$, where it is written as:
\vspace{-0.8em}
\begin{equation}
\vspace{-0.6em}
\widehat{\mathcal{F}}^{mp}_{k} \left( x, y \right) =
\left\{
\begin{array}{ll}
\Bar{\mathcal{F}}_{k \rightarrow k+1}, &\left( x, y \right) \in f_d \left( \widehat{\mathcal{M}}_k \right) \\
0, &otherwise
    \end{array} \right.
\end{equation}
Finally, we obtain $\widehat{\mathcal{F}}_{k \rightarrow k+1}$ of the $k$-th edited frame via:
\vspace{-0.5em}
\begin{equation} \label{eq: compose-flow}
\vspace{-0.6em}
\widehat{\mathcal{F}}_{k \rightarrow k+1} = \mathcal{F}_{k \rightarrow k+1} \odot \left( 1 - f_d \left( \widehat{\mathcal{M}}_k \right) \right) + \widehat{\mathcal{F}}^{mp}_{k}.
\end{equation}
Herein, $f_d \left( \cdot \right)$ and $\odot$ refer to the binary dilation and the Hadamard production operations, respectively, where we apply them on $\widehat{\mathcal{M}}_k$ to ensure that the simulated motion covers all regions of the edited contents.
Once $\widehat{\mathcal{F}}_{k \rightarrow k+1}$ is simulated, we obtain $\widehat{\mathcal{M}}_{k+1}$ via warping $\widehat{\mathcal{M}}_k$, written as:
\vspace{-0.4em}
\begin{equation} \label{eq: mask-warping}
\vspace{-0.6em}
    \widehat{\mathcal{M}}_{k+1} = f_{w} \left( \widehat{\mathcal{M}}_k, \widehat{\mathcal{F}}_{k \rightarrow k+1} \right),
\end{equation}
where $f_{w} \left( \cdot \right)$ denotes the warping operation.
By iteratively simulating the optical flows from $k=1$ to $k=N-1$, we eventually obtain the optical flows $\widehat{\mathcal{F}} = \{ \widehat{\mathcal{F}}_{1 \rightarrow 2}, \dots, \widehat{\mathcal{F}}_{N-1 \rightarrow N} \}$ of all edited frames.

\begin{figure}[t!]
\centering
\includegraphics[width=1.0\linewidth, trim=0 0 0 0]{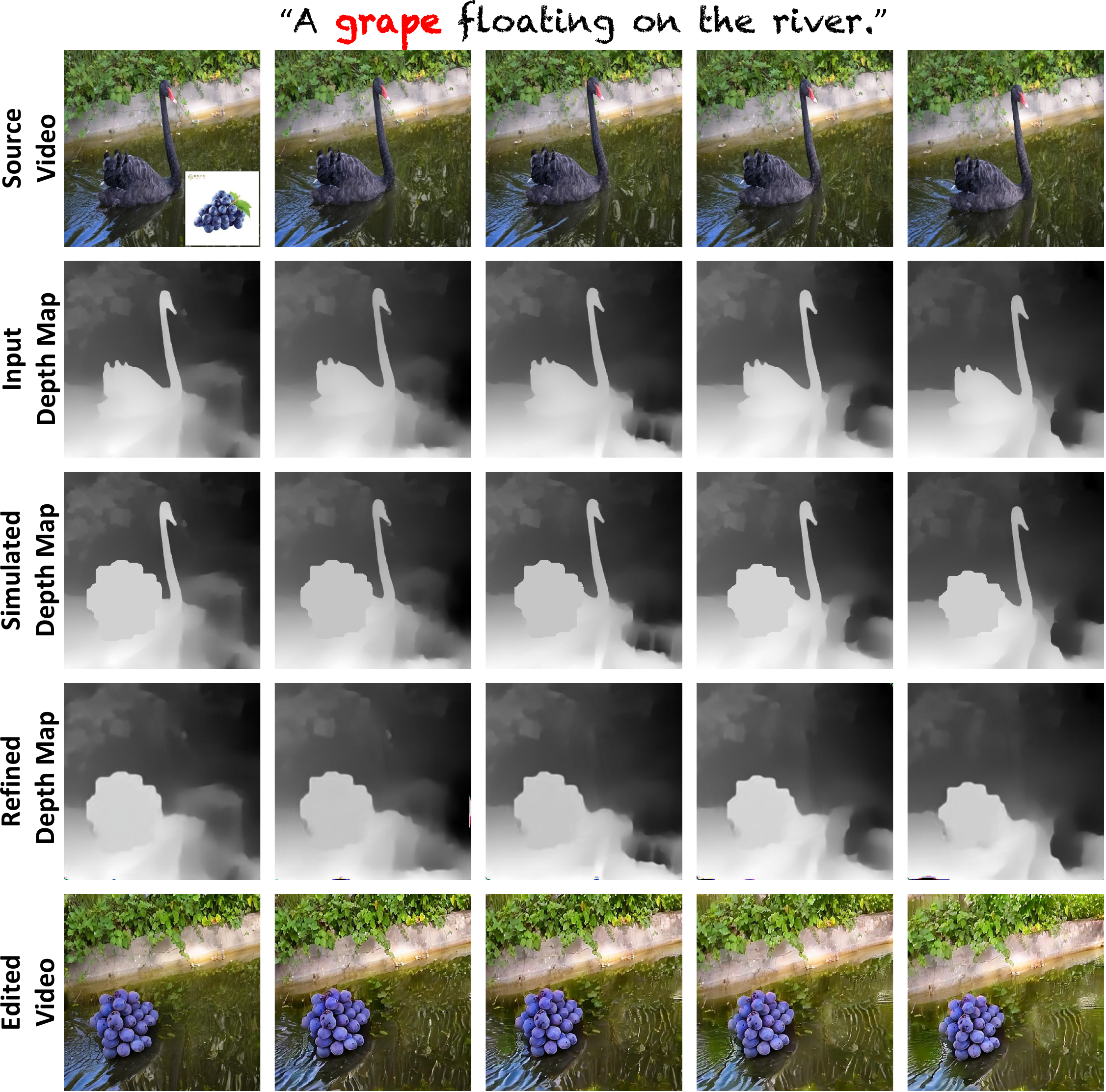}
\vspace{-1.8em}
\caption{
\textbf{Visualizations of the intermediate results by ISA.}
}
\vspace{-1.7em}
\label{fig: isa-visualziation}
\end{figure}

\begin{table*}[t!]
  \centering
  \setlength{\tabcolsep}{0.32em}
  \scalebox{1.02}{
  \begin{tabular}{lccccccccc}
   \toprule
   & \multicolumn{9}{c}{\textbf{\textsc{DAVIS-Edit-S} / \textsc{DAVIS-Edit-C}$_{\left( \Delta = \lvert \text{\textsc{C}} - \text{\textsc{S}} \rvert \right)}$}} \\
   \cmidrule{2-10} 
   \textbf{Method} & DOVER$^\uparrow$ & FVD$^\downarrow$ && WE$^\downarrow$ & CLIP-Temporal$^\uparrow$ && CLIP Score$^\uparrow$ && $\Bar{T}^\downarrow$ \\
   \midrule
   TokenFlow \cite{geyer-etal-2024-tokenflow} & \small66.36 / \small\underline{67.47}\textcolor{red}{$_{\text{(1.11)}}$} & \small17.33 / \small\underline{17.45}\textcolor{red}{$_{\text{(0.12)}}$} &&  \small18.58 / \small18.60\textcolor{red}{$_{\text{(0.02)}}$} & \small95.84 / \small\underline{95.61}\textcolor{red}{$_{\text{(0.23)}}$} && \small24.89 / \small24.12\textcolor{red}{$_{\text{(0.77)}}$} && \small5.81  \\
   FLATTEN \cite{cong-etal-2024-flatten} & \small63.86 / \small61.18\textcolor{red}{$_{\text{(2.68)}}$} & \small19.17 / \small21.65\textcolor{red}{$_{\text{(2.48)}}$} && \small17.29 / \small\underline{17.75}\textcolor{red}{$_{\text{(0.46)}}$} & \small95.39 / \small94.51\textcolor{red}{$_{\text{(0.88)}}$} && \small24.07 / \small23.24\textcolor{red}{$_{\text{(0.83)}}$} && \small4.23 \\
   Tune-A-Video \cite{jay-etal-2023-tuneavideo} & \small28.54 / \small34.63\textcolor{blue}{$_{\text{(6.09)}}$} & \small25.89 / \small26.76\textcolor{red}{$_{\text{(0.87)}}$} &&  \small89.63 / \small81.44\textcolor{blue}{$_{\text{(8.19)}}$} & \small91.82 / \small90.91\textcolor{red}{$_{\text{(0.91)}}$} && \small24.67 / \small\underline{24.89}\textcolor{blue}{$_{\text{(0.22)}}$} && \small20.23 \\ 
   Video-P2P \cite{liu-etal-2023-videop2p} & \small55.10 / \small51.22\textcolor{red}{$_{\text{(3.88)}}$} & \small17.22 / \small17.87\textcolor{red}{$_{\text{(0.65)}}$} &&  \small19.95 / \small18.82\textcolor{blue}{$_{\text{(1.13)}}$} & \small94.37 / \small93.51\textcolor{red}{$_{\text{(0.86)}}$} && \small24.72 / \small24.11\textcolor{red}{$_{\text{(0.61)}}$} && \small21.17 \\
   CoCoCo \cite{zi-etal-2024-cococo} & \small66.81 / \small66.12\textcolor{red}{$_{\text{(0.69)}}$} & \small18.13 / \small18.41\textcolor{red}{$_{\text{(0.28)}}$} &&  \small16.24 / \small18.47\textcolor{red}{$_{\text{(2.23)}}$} & \small\underline{96.07} / \small94.97\textcolor{red}{$_{\text{(1.10)}}$} && \small24.36 / \small23.24\textcolor{red}{$_{\text{(1.12)}}$} && \small\textbf{1.55} \\
   AnyV2V \cite{ku-etal-2024-anyv2v} & \small\underline{66.82} / \small65.01\textcolor{red}{$_{\text{(1.72)}}$} & \small\underline{14.87} / \small17.83\textcolor{red}{$_{\text{(2.96)}}$} &&  \small\textbf{15.35} / \small18.26\textcolor{red}{$_{\text{(2.91)}}$} & \small95.66 / \small94.36\textcolor{red}{$_{\text{(1.30)}}$} && \small\underline{25.09} / \small24.32\textcolor{red}{$_{\text{(0.77)}}$} && \small8.28 \\
   DMT \cite{danah-etal-2024-dmt} & \small59.27 / \small57.45\textcolor{red}{$_{\text{(1.82)}}$} & \small19.53 / \small21.64\textcolor{red}{$_{\text{(2.11)}}$} &&  \small16.65 / \small19.89\textcolor{red}{$_{\text{(3.24)}}$} & \small94.11 / \small93.58\textcolor{red}{$_{\text{(0.53)}}$} && \small24.91 / \small24.51\textcolor{red}{$_{\text{(0.40)}}$} && \small8.88 \\
   \rowcolor{lightgray}
   \textbf{\textsc{StableV2V}} & \small\textbf{67.78} / \small\textbf{70.80}\textcolor{blue}{$_{\text{(3.02)}}$} & \small\textbf{13.77} / \small\textbf{17.18}\textcolor{red}{$_{\text{(3.41)}}$} &&  \small\underline{15.95} / \small\textbf{15.27}\textcolor{blue}{$_{\text{(0.68)}}$} & \small\textbf{96.34} / \small\textbf{96.83}\textcolor{blue}{$_{\text{(0.49)}}$} && \small\textbf{25.46} / \small\textbf{25.68}\textcolor{blue}{$_{\text{(0.22)}}$} && \small\underline{3.14} \\
   \midrule
   AnyV2V \cite{ku-etal-2024-anyv2v} & \small65.83 / \small64.56\textcolor{red}{$_{\text{(1.27)}}$} & \small12.97 / \small15.25\textcolor{red}{$_{\text{(2.28)}}$} &&  \small24.47 / \small25.61\textcolor{red}{$_{\text{(1.14)}}$} & \small95.89 / \small96.13\textcolor{red}{$_{\text{(0.24)}}$} && \small25.41 / \small24.79\textcolor{red}{$_{\text{(0.62)}}$} && \small8.43 \\
   \rowcolor{lightgray}
   \textbf{\textsc{StableV2V}} & \small\textbf{67.58} / \small\textbf{68.42}\textcolor{blue}{$_{\text{(0.84)}}$} & \small\textbf{12.36} / \small\textbf{14.87}\textcolor{red}{$_{\text{(2.51)}}$} && \small\textbf{22.17} / \small\textbf{21.23}\textcolor{blue}{$_{\text{(0.94)}}$} & \small\textbf{96.51} / \small\textbf{96.71}\textcolor{blue}{$_{\text{(0.20)}}$} && \small\textbf{26.24} / \small\textbf{26.55}\textcolor{blue}{$_{\text{(0.31)}}$} && \small\textbf{3.23} \\ 
   \bottomrule
\end{tabular}
}
\vspace{-0.8em}
\caption{
\textbf{Quantitative results of \textsc{StableV2V} on text- (top) and image-based (bottom) evaluation settings of \textsc{Davis-Edit}}, compared to existing methods \cite{geyer-etal-2024-tokenflow, cong-etal-2024-flatten, 
jay-etal-2023-tuneavideo, liu-etal-2023-videop2p, zi-etal-2024-cococo, ku-etal-2024-anyv2v, danah-etal-2024-dmt} with respect to DOVER \cite{wu-etal-2023-dover}, FVD \cite{thomas-etal-2019-fvd}, Warping Error (WE), CLIP-Temporal \cite{peruzzo-etal-2024-vase}, CLIP scores \cite{hessel-etal-2021-clipscore}, and averaged inference time 
(termed $\Bar{T}$, in units of minutes), where the \textbf{best} and \underline{second best} results are \textbf{boldfaced} and \underline{underlined}.
Results on DOVER, FVD, WE, CLIP-Temporal, and CLIP scores are scaled by $10^{-2}$, $10^2$, $10^{-5}$, $10^{-2}$, and $10^{-2}$, respectively.
Herein, performance gain and drop by comparing \textsc{DAVIS-Edit-C} to \textsc{DAVIS-Edit-S} are highlighted in \textcolor{blue}{blue} and \textcolor{red}{red}, correspondingly.
}
\vspace{-1.6em}
\label{tab: quantitative-comparison}
\end{table*}

\noindent \textbf{Depth Simulation.}
Once we simulate the motion information of the edited video, the next step is to obtain the guidance for the image-to-video generator, i.e., the depth maps.
In doing so, we conduct procedures similar to that in motion simulation.
Specifically, we firstly adopt an off-the-shelf depth estimator (i.e., MiDaS \cite{rene-etal-2022-midas}) to extract the depth maps $\mathcal{D} = \{ \mathcal{D}_1, \dots, \mathcal{D}_{N} \}$ from $\mathcal{X}$.
Given the $k$-th ($k \sim \{ 1 \dots N \}$) depth map $\mathcal{D}_k$, we compute the average depth similar to the process of Eq. (\ref{eq: average-flow}), formulated by:
\vspace{-0.6em}
\begin{equation}
\vspace{-0.7em}
    \Bar{\mathcal{D}}_k = \frac{1}{\mathcal{M}_k} \sum_{\left( i, j \right) \in \mathcal{M}_k} \mathcal{D}_k \left( i, j \right),
\end{equation}
where $\left( i, j \right)$ represents the pixel at the $i$-th row and $j$-th column.
Then, we conduct the depth pasting operation on $\widehat{\mathcal{M}}_k$ to propagate the depth information, where the average depth $\widehat{\mathcal{D}}^{dp}_k \left( x, y \right) = \Bar{\mathcal{D}}_k$ if $\left( x, y \right) \in \widehat{\mathcal{M}}_k$ otherwise $0$.
Finally, we construct the $k$-th simulated depth map $\widehat{\mathcal{D}}_k$ via composing:
\vspace{-0.7em}
\begin{equation}
\vspace{-0.5em}
    \widehat{\mathcal{D}}_k = \mathcal{D}_k \odot \left( 1 - \widehat{\mathcal{M}}_k \right) + \widehat{\mathcal{D}}^{dp}_k.
\end{equation}
By iterating all depth maps $\mathcal{D} = \{ \mathcal{D}_1, \dots, \mathcal{D}_{N-1} \}$, we are able to obtain the simulated depth map $\widehat{\mathcal{D}} = \{ \widehat{\mathcal{D}}_1, \dots, \widehat{\mathcal{D}}_N \}$ of all edited video frames.
Since the simulated depth maps $\widehat{\mathcal{D}}$ are obtained via composing, we observe that $\widehat{\mathcal{D}}$ often contains unnecessary depth information in the regions of the original contents, as is shown in Fig. \ref{fig: isa-visualziation}, indicating that $\widehat{\mathcal{D}}$ needs to be further refined to ensure its preciseness.

\noindent \textbf{Shape-guided Depth Refinement.}
To refine $\widehat{\mathcal{D}}$, we draw inspirations from existing video inpainting methods \cite{zhou-etal-2023-propainter} that adopt completion networks to repair optical flows, and propose a depth refinement network based on such paradigm.\footnote{We illustrate the implementation details of the shape-guided depth refinement network in Sec. \ref{sec: shape-guided-depth-refinement-network} of our supplementary materials.}
Furthermore, we integrate the first-frame shape mask $\widehat{\mathcal{M}}_1$ into it to ensure the shape consistency of refinement.
Given $\mathcal{M}$ and $\widehat{\mathcal{M}}$, the mask regions $\mathcal{M}_r$ and the masked depth maps $\widehat{\mathcal{D}}_m$ are obtained through:
\vspace{-0.4em}
\begin{equation} \label{eq: prepare-refinement-inputs}
\vspace{-0.7em}
\begin{aligned}
    \mathcal{M}_r &= f_d \left( \left( 1 - \widehat{\mathcal{M}}  \right) \odot \mathcal{M} \right),
    \\
    \widehat{\mathcal{D}}_m &= \mathcal{M}_r \odot \widehat{\mathcal{D}}.
\end{aligned}
\end{equation}
Then, we send the concatenation of $\widehat{\mathcal{D}}_m$, $\mathcal{M}_r$, and $\widehat{{\mathcal{M}}}_1$ into the shape-guided refinement network $f_{r} \left( \cdot \right)$, resulting in the final depth maps $\widehat{\mathcal{D}}_r$, where the process is written as:
\vspace{-0.5em}
\begin{equation} \label{eq: depth-refinement}
\vspace{-0.5em}
    \widehat{\mathcal{D}}_r = f_{r} \left( \widehat{\mathcal{D}}_m, \mathcal{M}_r, \widehat{{\mathcal{M}}}_1 \right).
\end{equation}
In this way, ISA is able to obtain the accurately simulated depth maps $\widehat{\mathcal{D}}_r$ of the edited video, where $\widehat{\mathcal{D}}_r$ later play a pivotal role in offering precise guidance for CIG.

\subsection{Conditional Image-to-video Generator}
Once we obtain $\widehat{\mathcal{D}}_r$, the final goal of CIG is to generate the edited video $\widehat{\mathcal{X}}$.
Specifically, CIG consists of two components, i.e., the controller model and the image-to-video generator, where we use Ctrl-Adapter \cite{lin-etal-2024-ctrladapter} as a controller to inject $\widehat{\mathcal{D}}_r$, and leverage I2VGen-XL \cite{zhang-etal-2023-i2vgenxl} to propagate the edited contents from $\widehat{\mathcal{X}}_1$ to all other frames in $\widehat{\mathcal{X}}$, respectively.
Given the corresponding text prompt $\mathcal{P}_t$ and $\widehat{\mathcal{D}}_r$, CIG produces the final edited video $\widehat{\mathcal{X}}$ through:
\vspace{-0.5em}
\begin{equation} 
    \widehat{\mathcal{X}} = 
    \{ \widehat{\mathcal{X}}_1, \dots, \widehat{\mathcal{X}}_N \} =
    f_{CIG} \left( \widehat{{\mathcal{X}}}_1, \mathcal{P}_t, \mathcal{E}_c \left( \widehat{\mathcal{D}}_r \right) \right).
\end{equation}

\section{Experimental Settings}
\vspace{-0.3em}
In this section, we illustrate our experimental settings from aspects of evaluation setup, testing benchmark, baselines, and metrics, whose details are presented as follows.

\noindent \textbf{Evaluation Setup.}
In our experiments, we summarize and evaluate existing video editing studies based on two mainstream setups, i.e., \textit{text-} and \textit{image-based} evaluation.
For \textit{text-based} evaluation, we adopt captions with only their object words modified to generate the edited videos.
For \textit{image-based} evaluation, we utilize reference images as external prompts to produce the edited videos.

\begin{figure*}[t!]
\centering
\includegraphics[width=1.0\linewidth, trim=0 0 0 0]{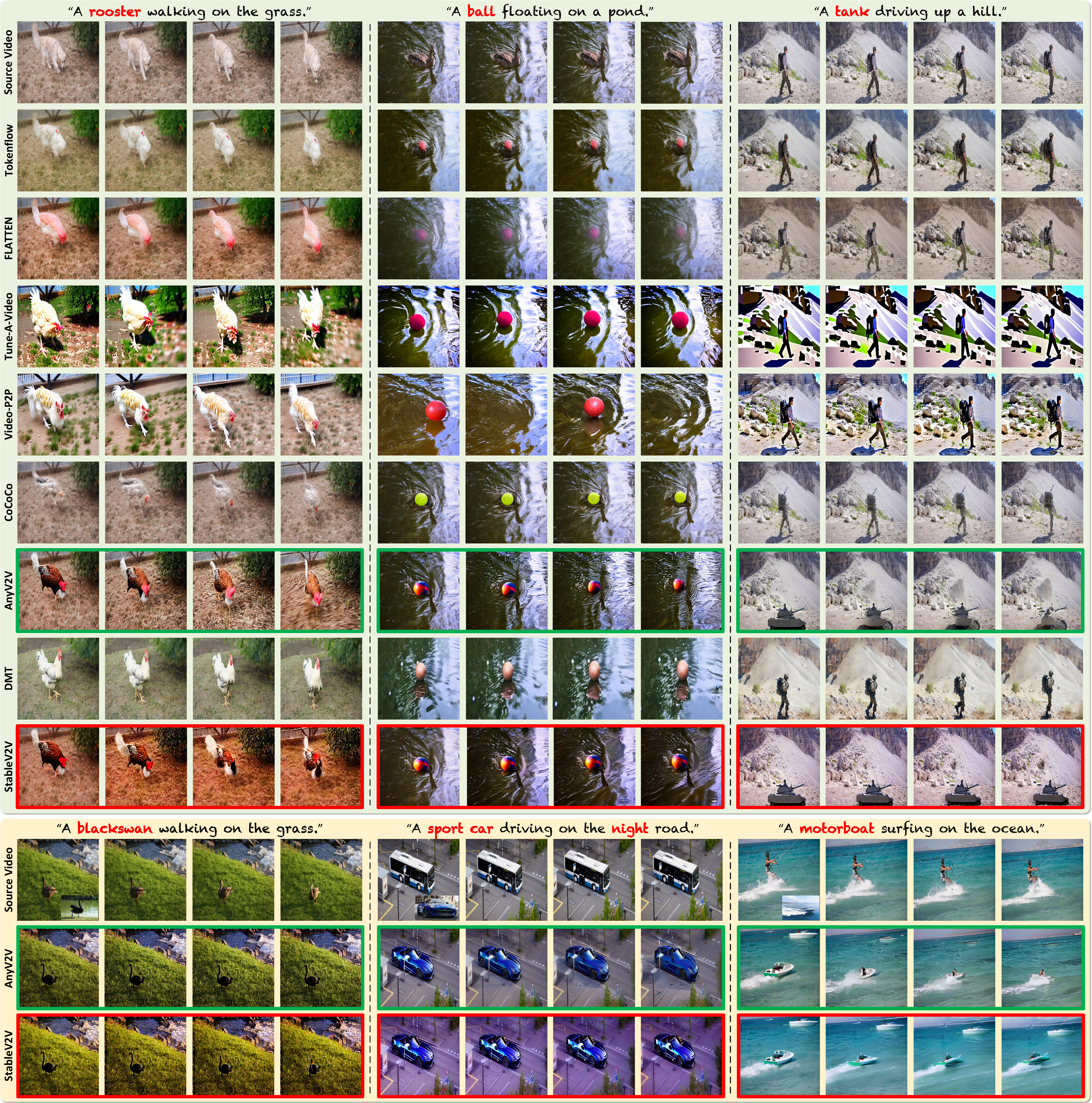}
\vspace{-2.1em}
\caption{
\textbf{Qualitative comparison of text- and image-based editing,} with their backgrounds highlighted in green and yellow, respectively.
Note that results of AnyV2V \cite{ku-etal-2024-anyv2v} (green bounding boxes) use the same first edited frames as ours (red bounding boxes).
}
\vspace{-1.5em}
\label{fig: qualitative-comparison}
\end{figure*}

\noindent \textbf{Testing Benchmark.}
For evaluation, we construct a testing benchmark, namely \textsc{DAVIS-Edit}, based on DAVIS \cite{ponttuset-etal-2018-davis}.
\textsc{DAVIS-Edit} contains two subsets \textsc{DAVIS-Edit-S} and \textsc{DAVIS-Edit-C}, which address the scenarios with similar (S) and changing (C) shapes, respectively.
Specifically, we select $26$ videos from DAVIS, and annotate the captions and images for them, obtaining $100$ cases eventually.\footnote{We illustrate more details of the proposed testing benchmark \textsc{DAVIS-Edit} in Sec. \ref{sec: implementation-details-of-davis-edit} of our supplementary materials.}

\noindent \textbf{Baselines.}
We compare \textsc{StableV2V} with several state-of-the-art video editing methods, including TokenFlow \cite{geyer-etal-2024-tokenflow}, FLATTEN \cite{cong-etal-2024-flatten}, Tune-A-Video \cite{jay-etal-2023-tuneavideo}, Video-P2P \cite{liu-etal-2023-videop2p}, CoCoCo \cite{zi-etal-2024-cococo}, AnyV2V \cite{ku-etal-2024-anyv2v}, and DMT \cite{danah-etal-2024-dmt}.
Notably, we use the same first edited frames in comparison with other first-frame-based methods such as AnyV2V.\footnote{Since we have no access to AVID \cite{zhang-etal-2024-avid}, VASE \cite{peruzzo-etal-2024-vase}, and I2VEdit \cite{ouyang-etal-2024-i2vedit}, we qualitatively compare \textsc{StableV2V} with them based on their demo videos, with details presented in Sec. \ref{sec: more-qualitative-comparison} of our supplementary materials.}

\noindent \textbf{Metrics.}
We evaluate all compared methods from four aspects, i.e., \textit{visual quality}, \textit{temporal consistency}, \textit{alignment}, and \textit{efficiency}.
For \textit{visual quality}, we utilize DOVER \cite{wu-etal-2023-dover} and FVD \cite{thomas-etal-2019-fvd} for evaluation.
For \textit{temporal consistency}, we compute the Warping Error (WE) of adjacent frames in the edited video, and adopt CLIP-Temporal following VASE \cite{peruzzo-etal-2024-vase}.
For \textit{alignment}, we leverage CLIP score \cite{hessel-etal-2021-clipscore} to measure the feature similarities of generated frames with the text prompts.
For \textit{efficiency}, we evaluate based on averaged inference time, where results are tested on the same A100 GPU with \texttt{torch.float16} precision.
Besides, we conduct user study to analyze with human evaluation.\footnote{In this paper, ``D.'', ``WE'', ``C.-T'', and ``C.S.'' denote the abbreviations of DOVER \cite{wu-etal-2023-dover}, Warping Error, CLIP-Temporal \cite{peruzzo-etal-2024-vase}, and CLIP score \cite{radford-etal-2021-clip} unless otherwise specified. Besides, DOVER, FVD, WE, CLIP-Temporal, and CLIP scores are scaled by $10^{-2}$, $10^2$, $10^{-5}$, $10^{-2}$, and $10^{-2}$.}\footnote{We recruit $17$ users, and show them with the inputs, prompts, and results, with $10$ and $11$ cases from \textsc{DAVIS-Edit-S} and \textsc{DAVIS-Edit-C}, respectively. Each user is asked to choose the videos with best quality without knowing the corresponding methods. Then, we compute the averaged top-1 preference percentage of all cases for comparison.}

\begin{table}[t!]
  \centering
  \setlength{\tabcolsep}{0.65em}
  \vspace{-0.5em}
  \scalebox{1.0}{\begin{tabular}{lccc}
   \toprule
   \textbf{Method} & \textbf{\textsc{D.-E.-S}} & \textbf{\textsc{D.-E.-C}} & \textbf{Avg.} \\
   \midrule
   TokenFlow \cite{geyer-etal-2024-tokenflow}   & 14.71\%   & 7.49\%    & 10.92\% \\
   FLATTEN \cite{cong-etal-2024-flatten}        & 3.53\%    & 1.60\%    & 2.52\% \\
   Tune-A-Video \cite{jay-etal-2023-tuneavideo} & 0.00\%    & 5.88\%    & 3.08\% \\
   Video-P2P \cite{liu-etal-2023-videop2p}      & 7.65\%    & 2.14\%    & 4.77\% \\
   CoCoCo \cite{zi-etal-2024-cococo}            & 10.58\%   & 8.56\%    & 9.52\% \\
   AnyV2V \cite{ku-etal-2024-anyv2v}            & 17.06\%   & \underline{23.53}\%   & 20.45\% \\
   DMT \cite{danah-etal-2024-dmt}               & \underline{21.18\%}   & \underline{23.53\%}   & \underline{22.41\%} \\
   \rowcolor{lightgray}
   \textbf{\textsc{StableV2V}}           & \textbf{25.29\%}      & \textbf{27.27\%} & \textbf{26.33\%} \\
   \bottomrule
\end{tabular}}
  \vspace{-0.8em}
  \caption{
  \textbf{Human evaluation results on \textsc{DAVIS-Edit-S} (``\textsc{D.-E.-S}'') and \textsc{DAVIS-Edit-C} (``\textsc{D.-E.-C}'').}
  }
\label{tab: human-evaluation}
\vspace{-2.2em}
\end{table}

\vspace{-0.6em}
\section{Results and Applications} \label{sec: results}
\vspace{-0.1em}

\noindent \textbf{Performance Comparison and Human Evaluation.}
Tab. \ref{tab: quantitative-comparison}, Fig. \ref{fig: qualitative-comparison}, and Tab. \ref{tab: human-evaluation} report the quantitative, qualitative comparisons, and human evaluation on \textsc{Davis-Edit}, respectively, compared to several existing methods \cite{geyer-etal-2024-tokenflow, cong-etal-2024-flatten, jay-etal-2023-tuneavideo, liu-etal-2023-videop2p, zi-etal-2024-cococo, danah-etal-2024-dmt, ku-etal-2024-anyv2v}.
Specifically, TokenFlow \cite{geyer-etal-2024-tokenflow} and FLATTEN \cite{cong-etal-2024-flatten} produce videos that are inconsistent with user prompts, and obtain inferior performance on most metrics, proving our motivation to address the shape inconsistency issue.
Similar trends are observed in Tune-A-Video \cite{jay-etal-2023-tuneavideo} and Video-P2P \cite{liu-etal-2023-videop2p}, with the video quality severely deteriorated, due to their incapabilities of modeling consistent motions with user prompts.
Although CoCoCo \cite{zi-etal-2024-cococo} and AnyV2V \cite{ku-etal-2024-anyv2v} improve the aforementioned methods to some extents, they struggle to handle challenging cases with significant shape change, especially when AnyV2V uses the same edited frame as ours, suggesting the deficiencies in these methods. 
DMT \cite{danah-etal-2024-dmt} is the most related study to ours, where it fails to follow the edited text prompts in some scenarios, and tends to produce contents with information loss in the backgrounds.
\textsc{StableV2V} consistently outperforms others with promising performance and video quality, where its results are also overwhelmingly preferred by users.
Notably, we observe that most methods obtain worse performance on \textsc{DAVIS-Edit-C}, whose cases comprise more complicated shape changes and are thus more challenging, however, \textsc{StableV2V} still obtains promising results and even gets improvements, owing to the fact that it ensures the consistency between the delivered motions and user prompts, thus will not be confused by misaligned motions when producing the final videos as others do.\footnote{We present more results in Sec. \ref{sec: more-qualitative-comparison} of our supplementary materials.}

\begin{figure}[t!]
\centering
\includegraphics[width=1.0\linewidth, trim=0 0 0 0]{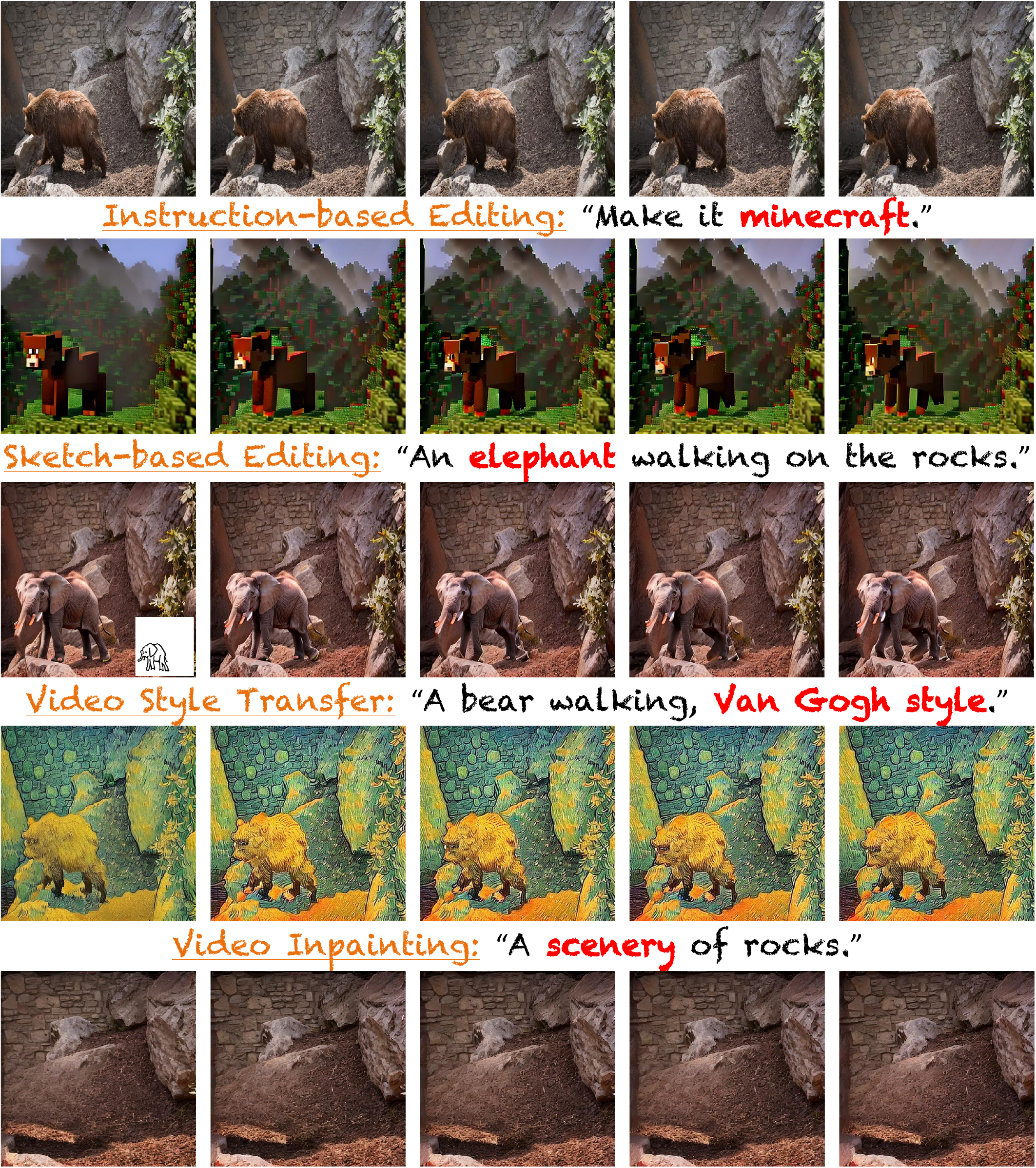}
\vspace{-2em}
\caption{
\textbf{More applications performed by \textsc{StableV2V}}, where the source video frames are shown in the first row.}
\vspace{-2em}
\label{fig: applications}
\end{figure}

\noindent \textbf{Efficiency Comparison.}
In our experiments, we observe that \textsc{StableV2V} demonstrates outstanding efficiency compared to other methods, as is reported in Tab. \ref{tab: quantitative-comparison}.
One can see that one-shot tuning-based methods \cite{jay-etal-2023-tuneavideo, liu-etal-2023-videop2p} take the most time (more than $20$ minutes) to edit a video due to their requirements of video-specific training, but the corresponding performance is not satisfying.
DDIM inversion-based methods \cite{geyer-etal-2024-tokenflow, ku-etal-2024-anyv2v, danah-etal-2024-dmt} also require massive time (around $6$ to $8$ minutes) to perform an complete editing process, where they need to prepare CNN features and attention maps via the inversion process.
FLATTEN \cite{cong-etal-2024-flatten} presents as an improved method that uses more efficient strategy to sample trajectories of optical flows, where \textsc{StableV2V} surpasses it with approximate $1.09$ minutes.
Eventually, CoCoCo serves as the best method in the comparison, however, it is worth noting that it also needs to train on Web10M \cite{webvid} for one epoch in advance, while \textsc{StableV2V} plays as a training-free solution for video editing.

\noindent \textbf{Applications.}
Despite of the aforementioned results, \textsc{StableV2V} also support other applications as is demonstrated in Fig. \ref{fig: applications}.
Herein, we adjust PFE according to the conducted application,
where \textsc{StableV2V} consistently handles different tasks, especially the ones that are susceptible to cause shape differences (e.g., instructions and sketches).
Notably in Fig.~\ref{fig: teasor} and \ref{fig: applications}, sketch-based editing offers a way for users to customize the shapes of edited contents, indicating the great potentials of applying \textsc{StableV2V} for real-world cases.
Notably, video inpainting represents an extreme scenario of shape differences in \textsc{StableV2V}, with the foreground object completely removed from the source video.
Particularly in ISA, $\widehat{\mathcal{M}}$ becomes all-zero maps since there is no foreground, and the pasting processes are subsequently skipped, where the shape-guided depth refinement network $f_r \left( \cdot \right)$ in such case aims to fully remove $\mathcal{D}$ and obtains depth maps of backgrounds to guide CIG.

\begin{figure}[t]
\centering
\includegraphics[width=1.0\linewidth, trim=0 0 0 0]{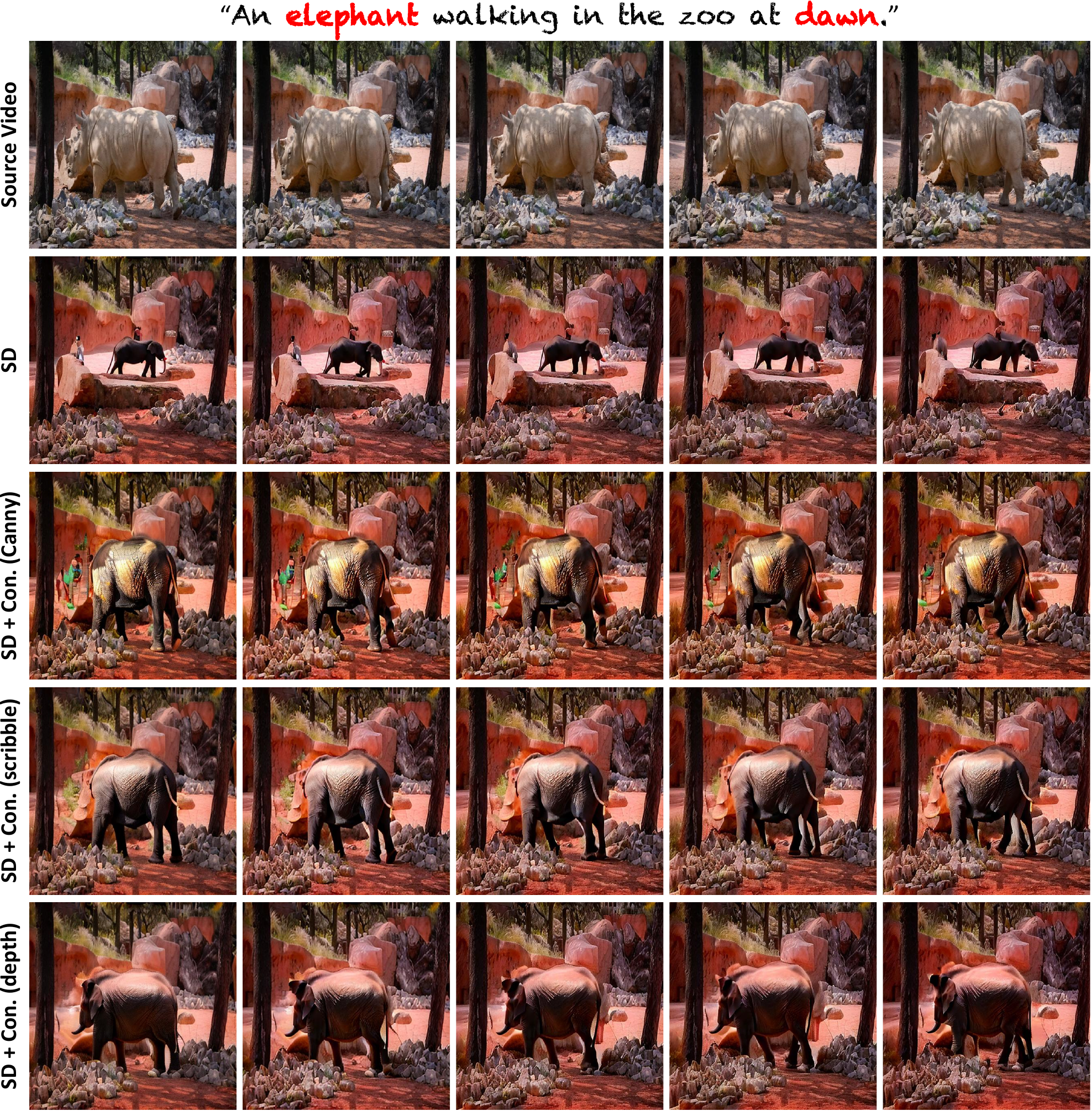}
\vspace{-2em}
\caption{
\textbf{Text-guided results under different settings of PFE.}
}
\vspace{-0.5em}
\label{fig: pfe-ablation}
\end{figure}

\vspace{-0.5em}
\section{Ablation Studies} \label{sec: ablation}
\vspace{-0.2em}
To further analyze \textsc{StableV2V}, we ablate its different components through conducting experiments under different settings of PFE and the depth simulation strategies, where details are presented in the following texts.

\begin{table}[t!]
  \centering
  \setlength{\tabcolsep}{0.3em}
  \vspace{-0.5em}
  \scalebox{1.0}{\begin{tabular}{lccccc}
   \toprule
   \textbf{Method} & \textbf{D.}$^\uparrow$ & \textbf{FVD}$^\downarrow$ & \textbf{WE}$^\downarrow$ & \textbf{C.-T}$^\uparrow$ & \textbf{C.S.}$^\uparrow$ \\
   \midrule
   SD  & 46.03 & 21.06 & 17.69 & 92.22 & 19.72 \\
   SD + Con. (Canny)  & 61.16 & 19.90 & 16.67 & 94.24 & 21.55 \\
   SD + Con. (scribble) & 64.08 & 14.70 & 16.69 & 95.66 & 24.75 \\
   \rowcolor{lightgray}
   \textbf{SD + Con. (depth)} & \textbf{67.78} & \textbf{13.77} & \textbf{15.95} & \textbf{96.34} & \textbf{25.46} \\
   \bottomrule
\end{tabular}}
\vspace{-0.8em}
\caption{\textbf{Evaluation scores under different settings of PFE}, evaluated on text-based editing of \textsc{DAVIS-Edit-S}.
}
\label{tab: effect-of-pfe}
\vspace{-2.2em}
\end{table}

\noindent \textbf{Effect of PFE on Text-based Editing.}
We evaluate the effect of PFE using various types of text-guided editors, with the corresponding results shown in Tab.~\ref{tab: effect-of-pfe} and Fig.~\ref{fig: pfe-ablation}.
Specifically, we use ``SD'' and ``SD + Con.'', referring to the SD inpaint model \cite{rombach-etal-2022-stable-diffusion} and the integrated framework that uses the ControlNet \cite{zhang-etal-2023-controlnet} to guide the inpainting process with conditions from the source video, respectively, where the condition types are illustrated in the parentheses.
We observe that ``SD'' often produces unstable edited contents like the ones in Fig. \ref{fig: pfe-ablation}, which later misguides the image-to-video generator, and produces video with inferior quality.
Using conditions significantly improves such limitation by enforcing the consistency, however, artifacts are observed due to the over-control by some conditions like Canny edge \cite{canny-etal-1986-canny}, with this situation alleviated in ``SD + Con. (scribble)'' and ``SD + Con. (depth)'' to some extents.
This experiments highlight the vitalness of the first edited frame, which offers superior flexibility on one hand, while on the other hand, it also determines how subsequent processes perform.

\begin{figure}[t]
\centering
\includegraphics[width=1.0\linewidth, trim=0 0 0 0]{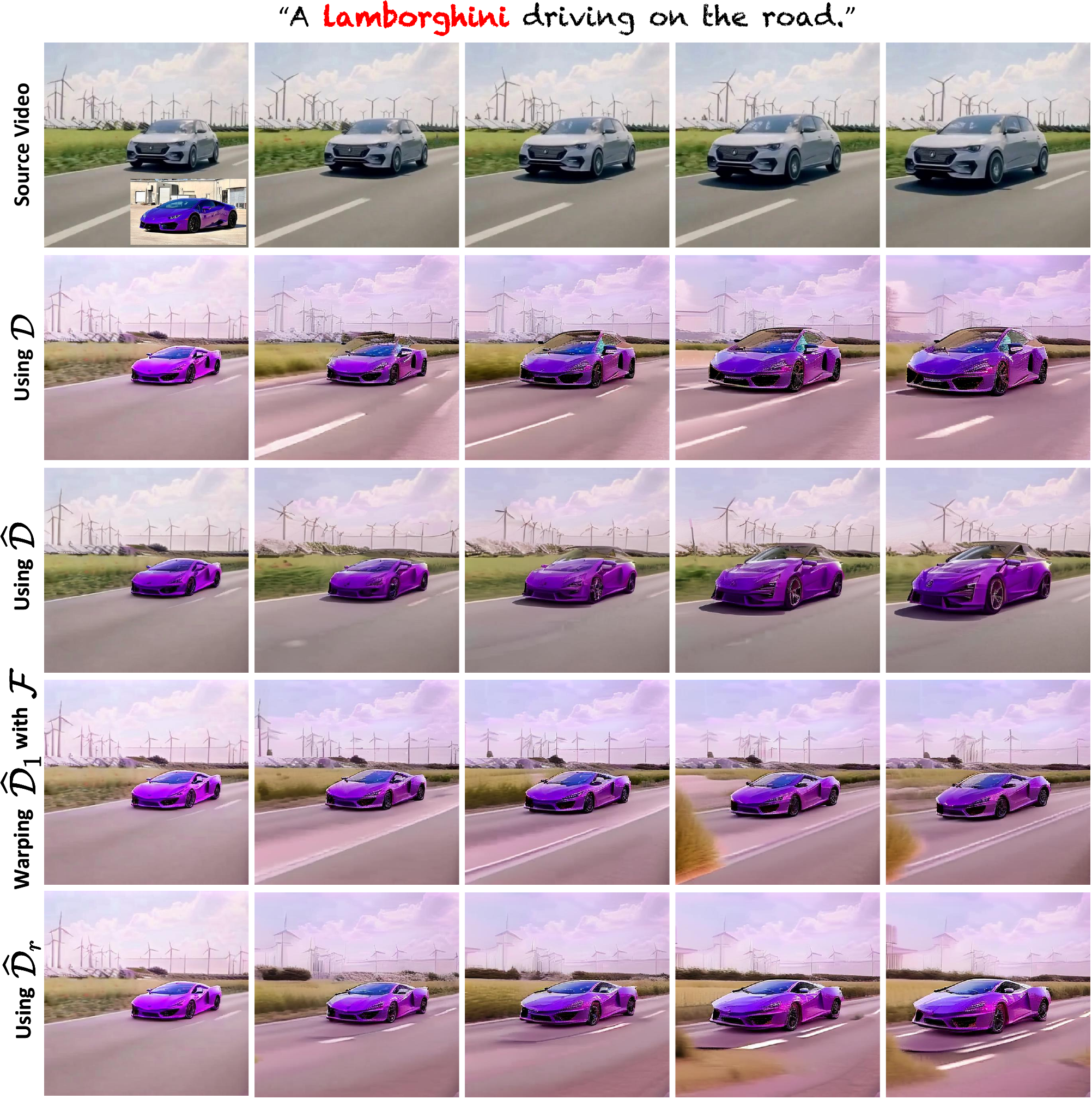}
\vspace{-2em}
\caption{
\textbf{Results under different depth simulation strategies.}
}
\vspace{-1em}
\label{fig: isa-ablation}
\end{figure}

\begin{table}[t!]
  \centering
  \setlength{\tabcolsep}{0.25em}
  \scalebox{1.0}{\begin{tabular}{lcccccc}
   \toprule
   \textbf{Depth Simulation} && \textbf{D.}$^\uparrow$ & \textbf{FVD}$^\downarrow$ & \textbf{WE}$^\downarrow$ & \textbf{C.-T}$^\uparrow$ & \textbf{C.S.}$^\uparrow$ \\
   \cmidrule{1-1} \cmidrule{3-7}
    Using $\mathcal{D}$ && 62.00 & 22.93 & 17.25 & 94.73 & 22.55 \\
    Using $\widehat{\mathcal{D}}$ && 66.46 & 16.62 & 16.36 & 95.94 & 24.55 \\
    Warping $\widehat{\mathcal{D}}_1$ with $\mathcal{F}$ && 64.54 & 19.14 & 16.83 & 95.33 & 23.71 \\
   \rowcolor{lightgray}
    \textbf{Using $\widehat{\mathcal{D}}_r$ (Ours)} && \textbf{67.78} & \textbf{13.77} & \textbf{15.95} & \textbf{96.34} & \textbf{25.46} \\
   \bottomrule
\end{tabular}}
\vspace{-0.7em}
  \caption{
  \textbf{Evaluation scores under different depth simulation strategies}, evaluated on text-based editing of \textsc{DAVIS-Edit-S}.
  }
\label{tab: effect-of-isa}
\vspace{-2em}
\end{table}

\noindent \textbf{Effect of the Depth Simulation Strategies.}
In \textsc{StableV2V}, depth map plays a vital role in transporting motions and guiding CIG, where we explore its effects via different simulation strategies, as is reported in Tab.~\ref{tab: effect-of-isa} and Fig.~\ref{fig: isa-ablation}.
Directly using $\mathcal{D}$ of source video suffers from issues similar to existing studies, where such depth maps misalign with the user prompts, so that incorrect motions are used for editing, thus leading to artifacts in results of CIG.
Similar results are shown when using $\widehat{\mathcal{D}}$ (w/o depth refinement), since $\widehat{\mathcal{D}}$ contain redundant regions like the ones in Fig.~\ref{fig: isa-visualziation}, indicating that depth refinement significantly boosts the accuracy of CIG guidance, thus ensuring that the edited video is consistent with user prompts.
Warping-based solution produces results with varying shapes due to the lack of motion pasting, where $\mathcal{F}$ fail to fully cover $\widehat{\mathcal{D}}_1$, especially when edited objects comprise larger sizes than the original ones, e.g., the case of editing a black swan to a bag in Fig.~\ref{fig: teasor}.

\vspace{-0.6em}
\section{Conclusion}
\vspace{-0.2em}
In this work, we present \textsc{StableV2V}, a shape-consistent video editing method that sequentially edits the first video frame, aligns the motions with user prompts, and finally produces the edited video with such consistent motions, with superior performance demonstrated on challenging applications.
Even so, \textsc{StableV2V} comprises several limitations due to the intrinsic problems of its paradigm, especially leading to potential working boundaries in cases with complicated motion patterns.
In our future work, we expect to address such issue, and propose an improved paradigm with more fine-grained motion modeling for video editing.\footnote{We analyze and discuss the limitations of our proposed method in Sec. \ref{sec: limitations} of our supplementary materials.}

{
    \small
    \bibliographystyle{ieeenat_fullname}
    \bibliography{reference}
}

\appendix

\twocolumn[{
 \renewcommand\twocolumn[1][]{#1}%
 \begin{center}
  \centering
  \includegraphics[width=1.0\textwidth]{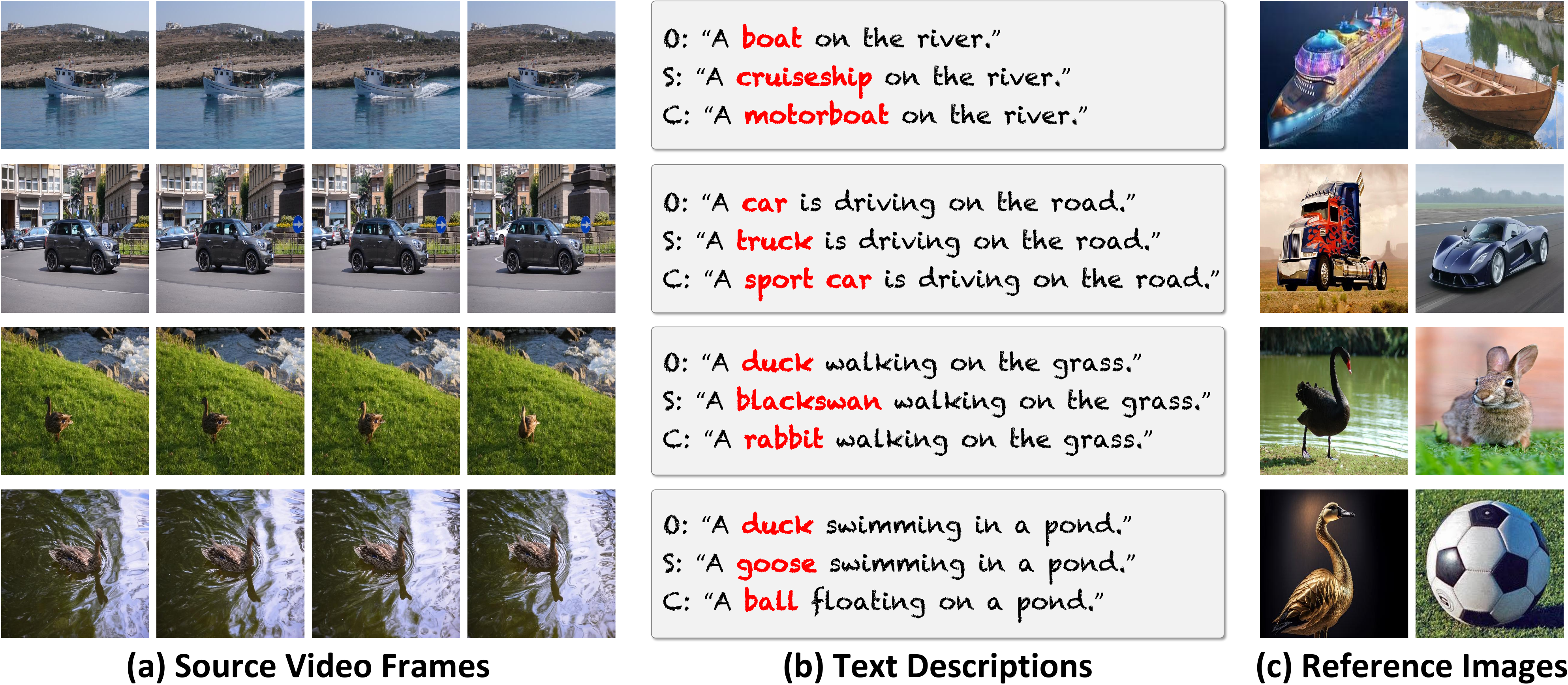}
  \vspace{-1.8em}
  \captionof{figure}{
  \textbf{Selected data samples and the corresponding annotations from \textsc{DAVIS-Edit},} with visualizations of (a) source video frames, (b) text descriptions, and (c) reference images highlighted in orange, green, and red, respectively.
  Specifically in (b), ``O'' represents the original (O) text description of the source video; ``S'' and ``C'' refers to the annotated captions indicating similar (S) and changing (C) shapes in the edited contents, respectively.
  Besides, we highlight the words depicting the main edited contents in red.
  In (c), we show the annotated images indicating similar and changing shapes on the left and right sides, respectively.
  }
  \label{fig: davis-edit-details}
  \vspace{-0.5em}
 \end{center}}]

\section*{Overview}
\vspace{-0.5em}

\begin{figure*}[t!]
  \centering
  \includegraphics[width=1.0\linewidth, trim=0 0 0 0]{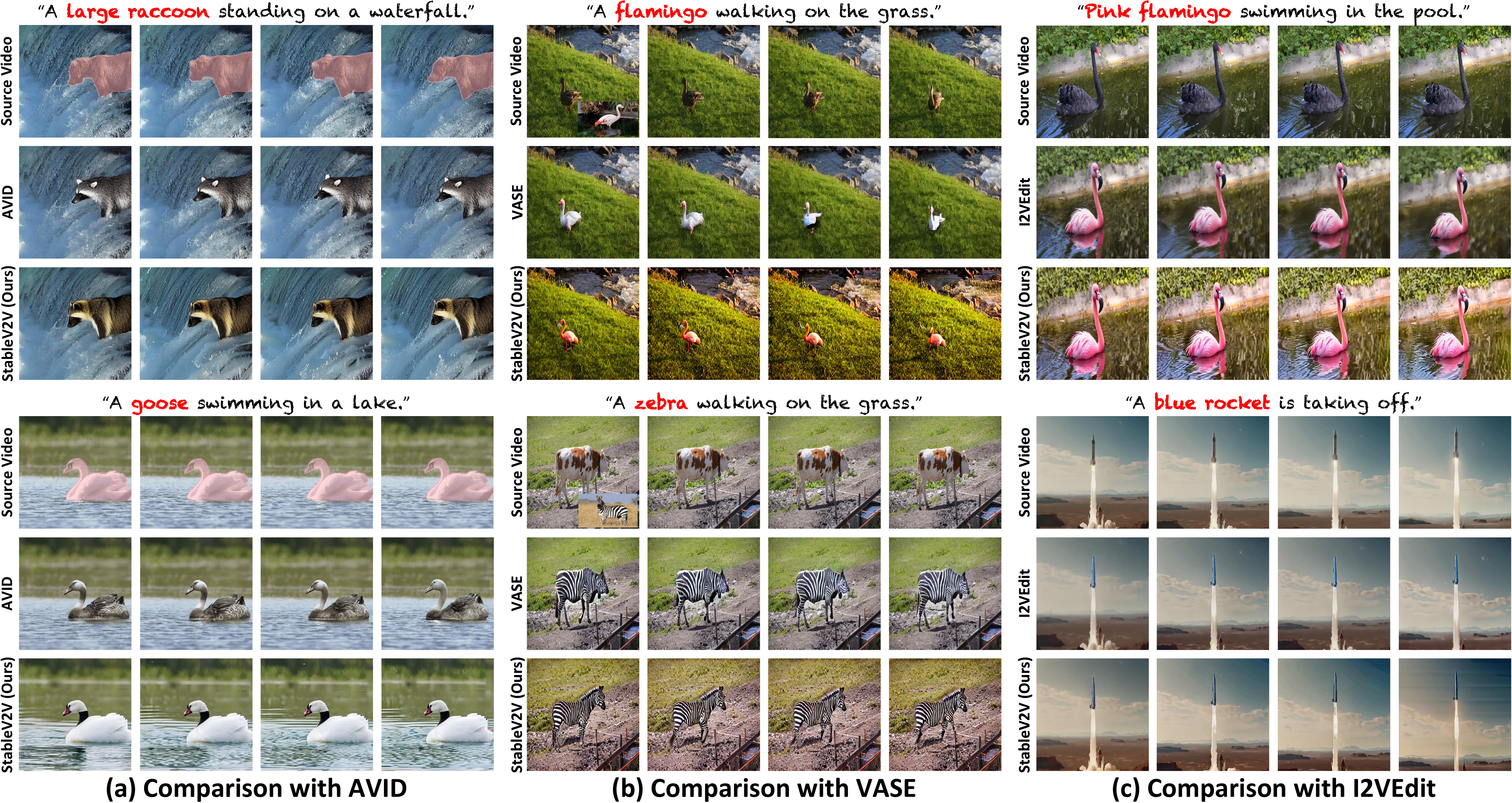}
  \vspace{-2em}
  \caption{
  \textbf{More qualitative comparison of \textsc{StableV2V},} compared to (a) AVID \cite{zhang-etal-2024-avid}, (b) VASE \cite{peruzzo-etal-2024-vase}, and (c) I2VEdit \cite{ouyang-etal-2024-i2vedit}.
  Note that we use the same first frame as the ones of I2VEdit \cite{ouyang-etal-2024-i2vedit} for comparison.
  }
  \vspace{-1em}
  \label{fig: more-qualitative-comparison}
  \end{figure*}

In our supplementary materials, we provide more details and results of \textsc{StableV2V}, so as to offer more insights into the proposed method, where we construct the contents following the structures below:
\begin{itemize}
    \item \textbf{Implementation Details of Shape-guided Depth Refinement Network.} The proposed depth refinement network plays a pivotal role in ensuring preciseness of depth guidance for \textsc{StableV2V}, where we illustrate its detailed implementation details from perspectives of the motivation, network architecture, and training in Sec. \ref{sec: shape-guided-depth-refinement-network}.
    \item \textbf{Implementation Details of the \textsc{DAVIS-Edit}.} \textsc{DAVIS-Edit} serves as the testing benchmark for the evaluation of \textsc{StableV2V}, where we report its implementation details in Sec. \ref{sec: implementation-details-of-davis-edit}, illustrating the annotation process of different prompts and some samples for demonstration.
    \item \textbf{More Qualitative Comparison.} In Sec. \ref{sec: more-qualitative-comparison}, we conduct the qualitative comparison with more video editing methods, especially the ones that are not open-sourced, including AVID \cite{zhang-etal-2024-avid}, VASE \cite{peruzzo-etal-2024-vase}, and I2VEdit \cite{ouyang-etal-2024-i2vedit}.
    \item \textbf{More Results.} In Sec. \ref{sec: more-results}, we demonstrate more qualitative results generated by \textsc{StableV2V}, from aspects of text-, image-based editing, and applications.
    \item \textbf{Limitations.} Although \textsc{StableV2V} achieves promising performance in various editing tasks, it also comprises several limitations due to its inherent problems, i.e., the paradigm based on pre-trained models and depth maps, where we discuss its working boundaries in Sec. \ref{sec: limitations}.
\end{itemize}
Notably, we offer the video format of all results (both main paper and this document) at \url{https://alonzoleeeooo.github.io/StableV2V}, and highly encourage readers to refer to them for a more intuitive experience of \textsc{StableV2V}.

\begin{figure*}[t!]
    \centering
    \includegraphics[width=1.0\linewidth, trim=0 0 0 0]{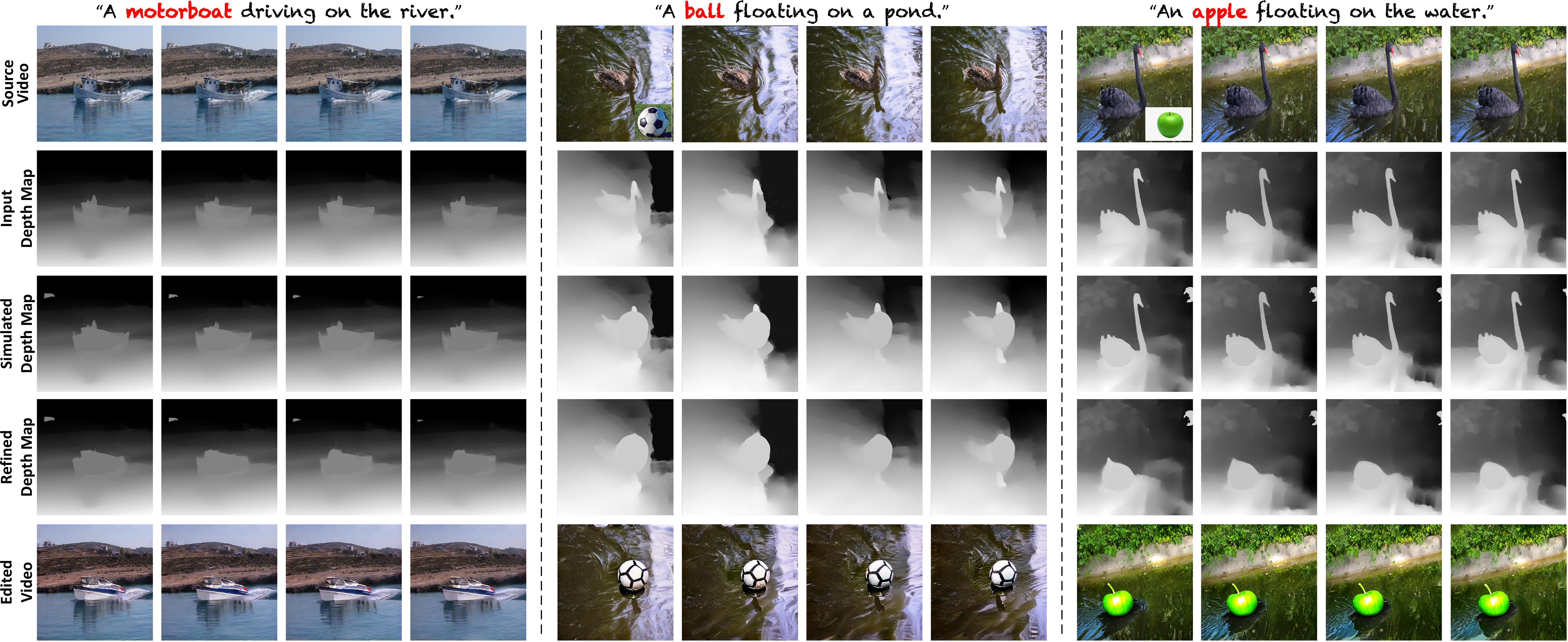}
    \vspace{-1.8em}
    \caption{
    \textbf{More visualizations of intermediate results in ISA}, where we show the reference images at the right-bottom corners.
    }
    \vspace{-1.5em}
    \label{fig: more-isa-results}
    \end{figure*}

\section{Implementation Details of Shape-guided Depth Refinement Network} \label{sec: shape-guided-depth-refinement-network}
In this section, we introduce the implementation details of shape-guided depth refinement network from various aspects, including its motivation, network architecture, and training details, as is presented in the following texts.

\noindent \textbf{Motivation and Network Architecture.}
The depth refinement network serves as a pivotal component in \textsc{StableV2V}, where it is highly associated with the preciseness of depth guidance for CIG, thus subsequently affecting the consisetncy of the edited video.
The final goal of such network is to calibrate the input depth maps by removing its redundant regions, meanwhile ensuring the consistency of the refined depth map with the corresponding edited first frame.
To build such network, we draw inspirations from the task of video inpainting \cite{zhou-etal-2023-propainter}, where optical flows, similar to the depth maps in \textsc{StableV2V}, normally serve as a pivotal guidance for the inpainting process.
Recently, VASE \cite{peruzzo-etal-2024-vase} borrows the same network architecture of the flow completion network in ProPainter \cite{zhou-etal-2023-propainter}, and adds an additional channel to the input layer to integrate the shape guidance, where the resulting network is used to offer flow guidance for reference-guided video editing.
Enlightened by the aforementioned studies, we adopt the same architecture as VASE does, and utilize the segmentation mask of the first edited frame as guidance for the refinement process.

\begin{figure}[t!]
  \centering
  \includegraphics[width=1.0\linewidth, trim=0 0 0 0]{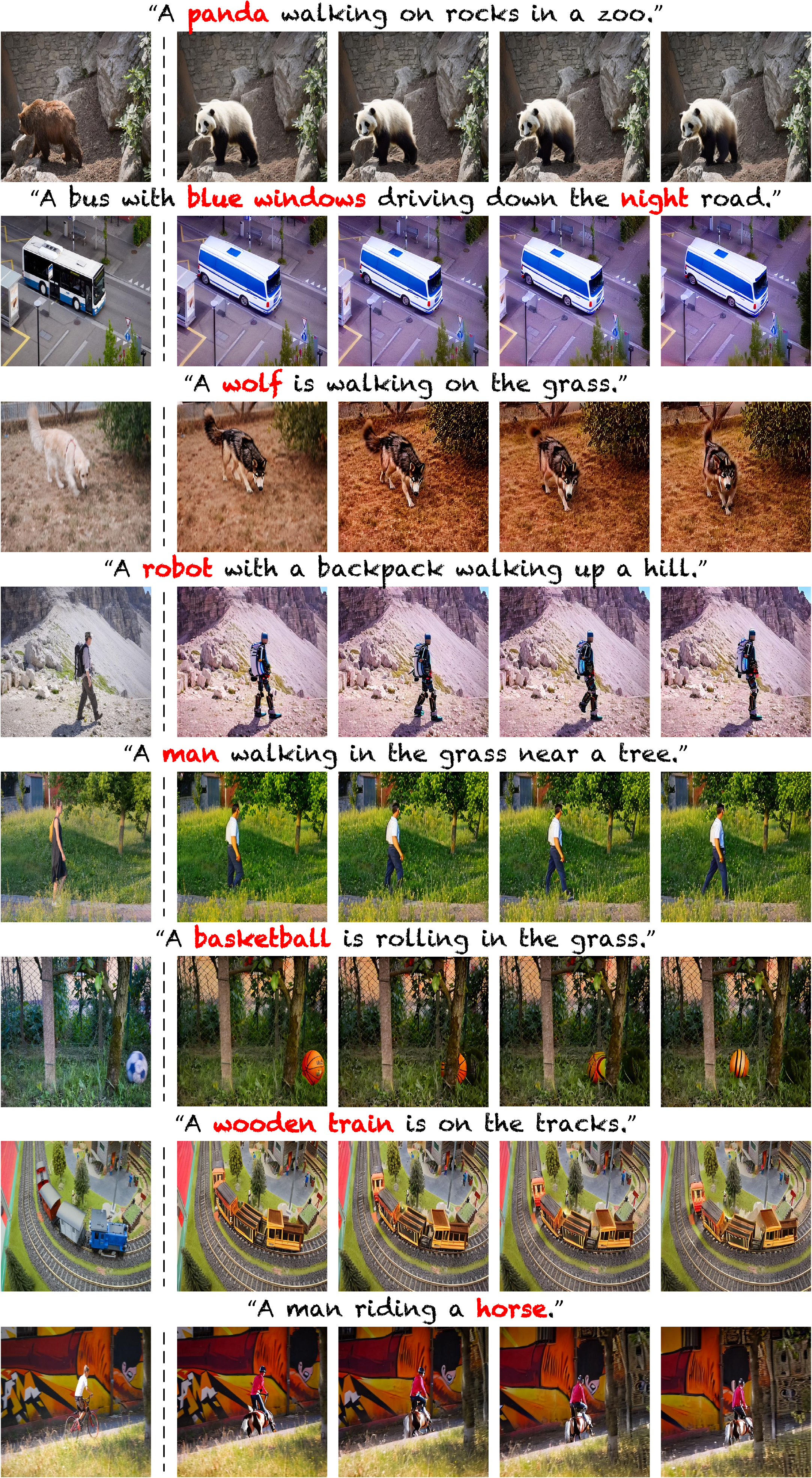}
  \vspace{-1.8em}
  \caption{
  \textbf{More text-based results generated by \textsc{StableV2V},} where we show the first frame of the source video in the first row.
  }
  \vspace{-1.5em}
  \label{fig: more-text-based-results}
  \end{figure}

\begin{figure}[t!]
  \centering
  \includegraphics[width=1.0\linewidth, trim=0 0 0 0]{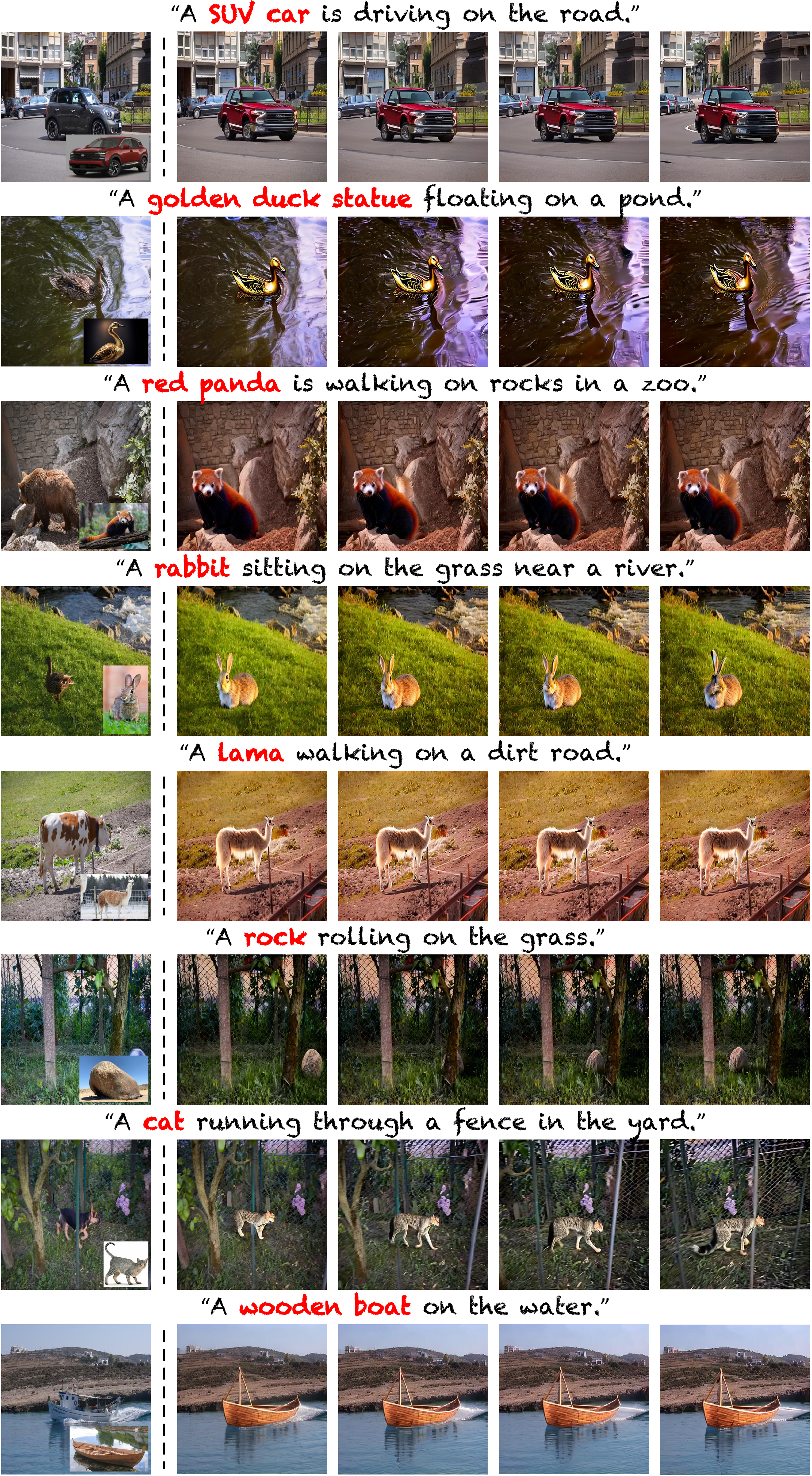}
  \vspace{-1.8em}
  \caption{
  \textbf{More image-based results generated by \textsc{StableV2V},} where we show the first frame of the source video in the first row for simplicity.
  Note that reference images are shown at the right-bottom corners of the first row.
  }
  \vspace{-1.5em}
  \label{fig: more-image-based-results}
  \end{figure}

\noindent \textbf{Training Details.}
We train the shape-guided depth refinement network on YouTube-VOS \cite{youtube-vos} dataset, whose training set consists of $3,471$ videos and the corresponding mask annotations in total.
To obtain the depth maps of all videos, we use an off-the-shelf depth estimator, i.e., MiDaS \cite{rene-etal-2022-midas}, to automatically annotate depth maps for all videos.
Once the data are pre-processed, we train the shape-guided depth refinement network for $50,000$ iterations, along with a batch size of $8$.
Specifically in each training step, we randomly sample $10$ frames of depth maps, and adopt the random mask generation algorithm in Flow-guided Transformer \cite{zhang-etal-2022-flowguided}.
We use AdamW \cite{adam} optimizer to update the model parameters, with the learning rate set to $0.99$.

\begin{figure*}[t!]
  \centering
  \includegraphics[width=1.0\linewidth, trim=0 0 0 0]{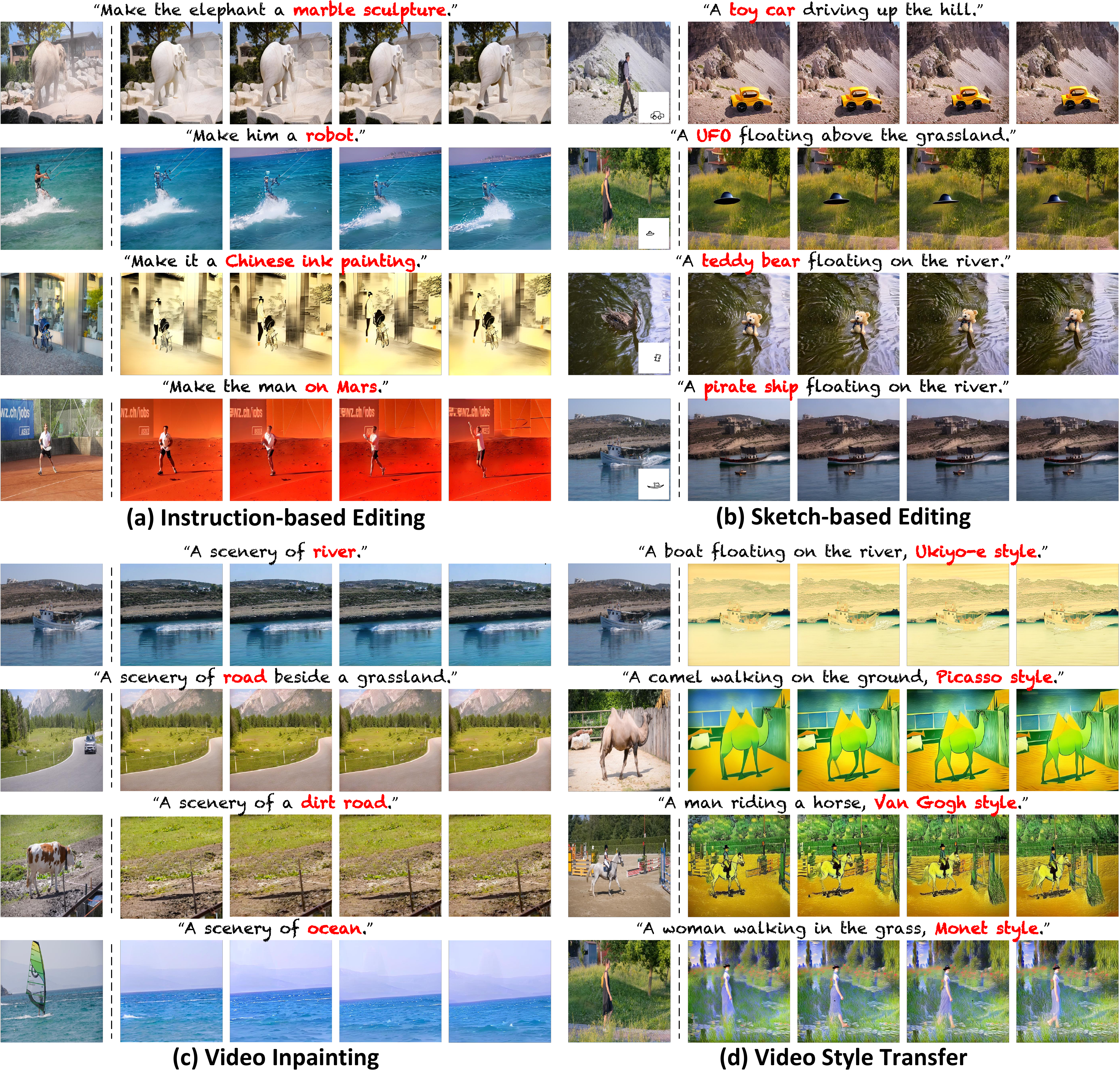}
  \vspace{-1.5em}
  \caption{
  \textbf{More results of applications conducted by \textsc{StableV2V},} including instruction-based editing, sketch-based editing, video style transfer, and video inpainting, whose backgrounds are highlighted in green, blue, red and yellow, respectively.
  }
  \vspace{-1.5em}
  \label{fig: more-applications}
  \end{figure*}

\vspace{-0.2em}
\section{Implementation Details of the \textsc{DAVIS-Edit}} \label{sec: implementation-details-of-davis-edit}
In this section, we illustrate more implementation details of our testing benchmark \textsc{DAVIS-Edit}.
\textsc{DAVIS-Edit} plays a crucial role in evaluating the performance of \textsc{StableV2V}, where we curate this testing benchmark to offer a standard to promote further studies in addressing the shape misalignment problem for video editing.
Fig. \ref{fig: davis-edit-details} demonstrates some samples selected from \textsc{DAVIS-Edit}, along with the example text prompts and reference images that we manually annotate.
To obtain the text prompts, we only modify specific words that describe the main elements of videos, e.g., objects and foregrounds, and put emphasis on embodying the shape difference problem during annotation.
For example, we use ``duck'' to replace ``blackswan'' to represent the setting with similar shapes of edited contents, and edit ``duck'' into ``rabbit'' for the scenario with changing shape.
For the annotation of reference images, we follow the similar principles, considering the variety of shape differences.
On top of that, we focus on collecting reference images that are tough for texts to illustrate, e.g., the Transformer truck in Fig. \ref{fig: davis-edit-details}, so as to highlight the impacts of image guidance in such setting.

\vspace{-0.6em}
\section{More Qualitative Comparison} \label{sec: more-qualitative-comparison}
\vspace{-0.3em}
In this section, we showcase more qualitative comparison with more methods, especially the ones that are not open-sourced yet, including AVID \cite{zhang-etal-2024-avid}, VASE \cite{peruzzo-etal-2024-vase}, and I2VEdit \cite{ouyang-etal-2024-i2vedit}.
Specifically, both AVID and VASE serve as learning-based solution for video editing, where AVID is a text-guided video inpainting framework initialized from SD Inpaint \cite{rombach-etal-2022-stable-diffusion}; VASE is fine-tuned based on a image-guided editor, i.e., Paint-by-Example \cite{yang-etal-2023-paint}, and mainly puts emphasis on object-centric video editing.
I2VEdit serves as a first-frame-based video editing method that trains video-specific LoRA \cite{lora} to model the motion patterns of the source video.
Since we do not have access to their code and model weights, we mainly compare \textsc{StableV2V} to their released demo video, with results presented in Fig. \ref{fig: more-qualitative-comparison}.
For fair comparison, we use the same reference images provided by VASE \cite{peruzzo-etal-2024-vase} in their demonstrated videos, and adopt the same first frame as the ones of I2VEdit \cite{ouyang-etal-2024-i2vedit}.

\noindent \textbf{Analyses.}
By comparing \textsc{StableV2V} to learning-based methods, i.e., AVID \cite{zhang-etal-2024-avid} and VASE \cite{peruzzo-etal-2024-vase}, it is observed that AVID \cite{zhang-etal-2024-avid} has possibilities in producing results with inconsistent textures, e.g., the case of editing a swan into a duck, suggesting its deficiencies in maintaining the temporal consistency.
VASE \cite{peruzzo-etal-2024-vase} produces results that merely transfer the textures of reference images into the edited videos, e.g., the cow-shape zebra in Fig. \ref{fig: more-qualitative-comparison}, since it is highly restricted by the input masks used in its inpainting paradigm.
The aforementioned results illustrate the typical issues in learning-based methods, where they are limited to editing scenarios with little shape changes due to the inpainting paradigm of their foundation models, i.e., SD Inpaint and Paint-by-Example, where such issues are significantly alleviated in \textsc{StableV2V}, since our first-frame-based scheme offers more flexiblity.
By comparing \textsc{StableV2V} to other first-frame-based method, i.e., I2VEdit \cite{ouyang-etal-2024-i2vedit}, two limitations are observed, where I2VEdit either produces results with information loss in the backgrounds, e.g., the case of editing the blackswan into a flamingo, or generates edited contents with simple motions like the case of a rising rocket.
Conversely, results generated by \textsc{StableV2V} comprise more detailed textures such as the waves on the river and the smoke emitted by the rocket, indicating that \textsc{StableV2V} not only offers robust consistency in the edited videos, but also ensures its video quality in details.

\begin{figure}[t]
  \centering
  \includegraphics[width=1.0\linewidth, trim=0 0 0 0]{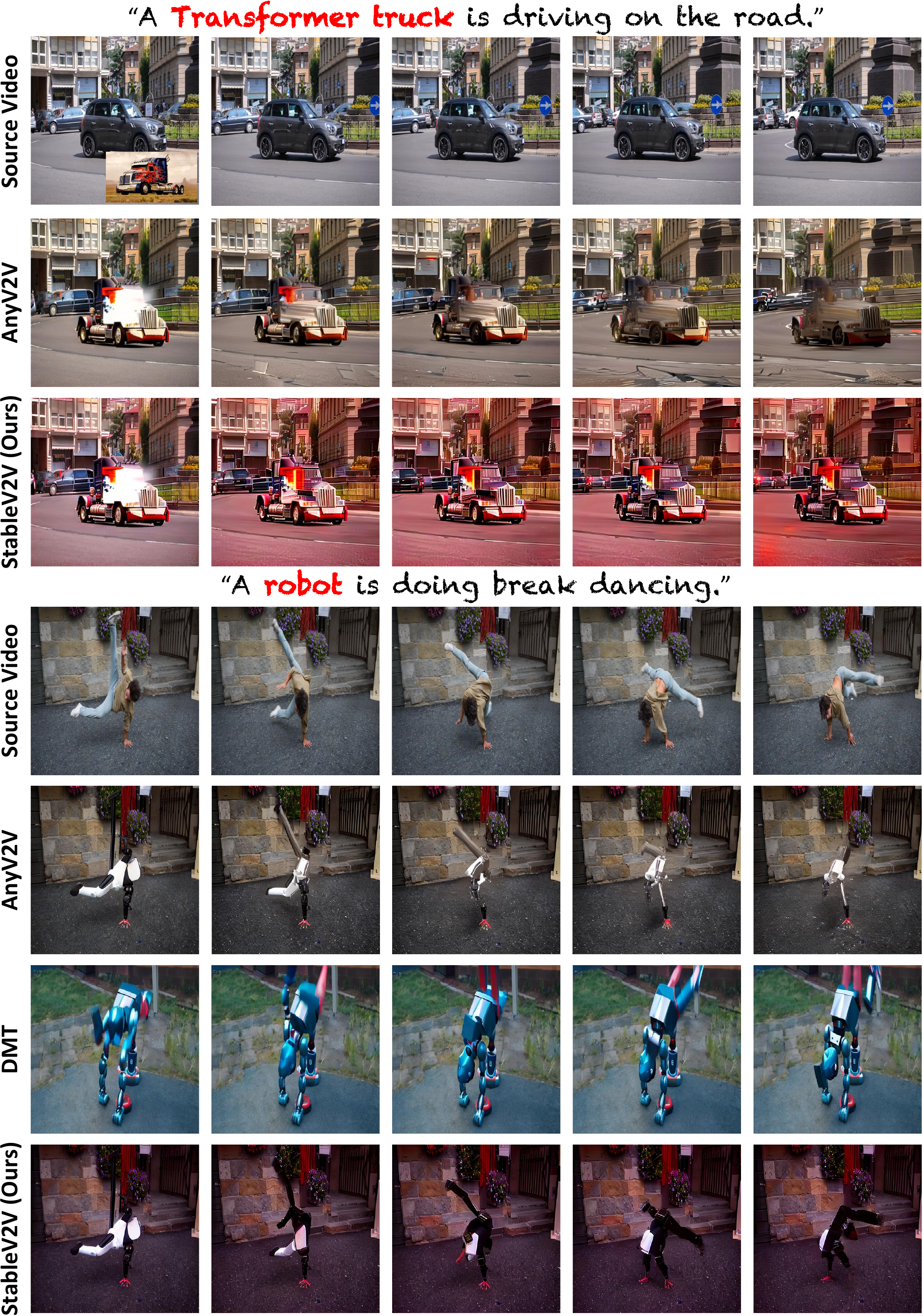}
  \vspace{-2em}
  \caption{
  \textbf{Failure cases of \textsc{StableV2V} illustrating the limitations of inherent problems of pre-trained models (top) and complicated motion patterns (bottom).}
  }
  \vspace{-1.8em}
  \label{fig: limitations}
  \end{figure}

\vspace{-0.6em}
\section{More Results} \label{sec: more-results}
\vspace{-0.3em}
In this section, we illustrate more results generated by \textsc{StableV2V}.
Specifically, we offer more visualizations of the intermediate results of ISA in Fig. \ref{fig: more-isa-results}.
Besides, we show several results on text- and image-based editing scenarios in Fig. \ref{fig: more-text-based-results} and Fig. \ref{fig: more-image-based-results}, respectively.
Also, we present more applications performed by \textsc{StableV2V} in Fig. \ref{fig: more-applications}.

\vspace{-0.5em}
\section{Limitations} \label{sec: limitations}
\vspace{-0.2em}
Although outperforming performance and applications are demonstrated by \textsc{StableV2V}, we observe that our proposed method also comprises several limitations due to the inherent problems that are caused by its paradigm, especially leading to potential working boundaries in cases that contain complicated motion patterns.
Therefore in this section, we analyze the limitations and working boundaries of \textsc{StableV2V}, with some failure cases shown in Fig. \ref{fig: limitations}, and discuss several potential solutions.
Details of the aforementioned analyses are illustrated in the following texts.

\noindent \textbf{inherent Problems of Pre-trained Models.}
Since \textsc{StableV2V} presents a training-free solution in addressing the misalignment problem between the motion controls and the edited contents, it relies on the use of pre-trained models and also suffers from severl inherent problems of them.
Specifically, this limitation occurs mostly in two components, i.e., PFE and CIG, where the former normally leverages off-the-shelf image editing methods; the latter is mainly designed based on a conditional generation paradigm for image-to-video generation, i.e., Ctrl-Adapter \cite{lin-etal-2024-ctrladapter}, since few studies are available in the existing literature.
For PFE, as is analyzed in Sec. \ref{sec: ablation} in our main paper, it comprises a certain degree of randomness in some text-guided editors such as SD Inpaint \cite{rombach-etal-2022-stable-diffusion}, where edited contents with undesired orientations might be produced, and then subsequently mis-guide the CIG module to produce inferior results.
For CIG, we observe that Ctrl-Adapter might lead to slight color discrepancy in several cases, especially when the edited contents are biased to certain colors, e.g., the case of editing the car into a Transformer truck in Fig. \ref{fig: limitations}.
Such color bias might be caused by the limited diversity and quality in the training data of Ctrl-Adapter, since its fine-tuning process may not require as much data as that used for its foundation model, i.e., I2VGen-XL \cite{zhang-etal-2023-i2vgenxl}.
Meanwhile, we observe that the generated textures are much more consistent than other studies, especially compared to the ones that also leverage I2VGen-XL, e.g., AnyV2V \cite{ku-etal-2024-anyv2v}, since ISA ensures the alignment between the edited contents and the delivered motions to CIG.
This finding indicates a potential solution to the above issue by considering ISA as a plug-and-play plugin, where we can integrate it into more powerful methods in the future once available.

\noindent \textbf{Working Boundaries in Complicated Motion Patterns.}
Another problem that \textsc{StableV2V} might suffer from is its limited capabilities in modeling motion patterns that are too complicated, e.g., the case of a man doing break dancing in Fig. \ref{fig: limitations}.
Similar results are observed in other studies like DMT \cite{danah-etal-2024-dmt} and AnyV2V \cite{ku-etal-2024-anyv2v}, where it is also tough for these methods to produce consistent results.
Such scenario serves as the challenging case that most existing methods struggle to handle, where the task of modeling fine-grained motions for video editing deserves studying in future works.

\end{document}